\documentclass[10pt,twocolumn,letterpaper]{article}

\usepackage{cvpr}
\usepackage{times}
\usepackage{epsfig}
\usepackage{graphicx}
\usepackage{amsmath}
\usepackage{amssymb}
\usepackage{float}

\usepackage[pagebackref=true,breaklinks=true,letterpaper=true,colorlinks,bookmarks=false]{hyperref}

\cvprfinalcopy 

\def\cvprPaperID{104} 

\ifcvprfinal\fi

\usepackage{times}
\usepackage{epsfig}
\usepackage{grffile}
\usepackage{amsmath}
\usepackage{amsfonts}
\usepackage{subfigure}
\usepackage{nonfloat}
\usepackage{url}
\usepackage{}
\usepackage{textcomp} 
\usepackage{color}
\usepackage{rotating}
\usepackage[export]{adjustbox}
\usepackage[table]{xcolor}

\usepackage{xspace}

\definecolor{red}{rgb}{0.95,0.4,0.4}
\definecolor{blue}{rgb}{0.4,0.4,0.95}
\definecolor{darkblue}{rgb}{0,0,0.8}
\definecolor{darkred}{rgb}{0.8,0,0}
\definecolor{darkgreen}{rgb}{0,0.5,0}
\definecolor{grey}{rgb}{0.6,0.6,0.6}

\definecolor{col1}{RGB}{213,229,255}
\definecolor{col2}{RGB}{128,179,255}

\newcommand*{\ea}{et al.\@\xspace}

\graphicspath{{ims/}{ims/kitti_eigen/}{ims/stereo_disp/}}

\usepackage[subtle, title=normal, bibliography=normal]{savetrees}

\usepackage{pdfpages}

\begin{document}
\title{Unsupervised Monocular Depth Estimation with Left-Right Consistency}

\author{Cl\'{e}ment Godard\hspace{40pt}Oisin Mac Aodha\hspace{40pt}Gabriel J. Brostow\\
University College London\\
http://visual.cs.ucl.ac.uk/pubs/monoDepth/
}

\maketitle

%
%
\begin{abstract}
Learning based methods have shown very promising results for the task of depth estimation in single images.
However, most existing approaches treat depth prediction as a supervised regression problem and as a result, require vast quantities of corresponding ground truth depth data for training.
Just recording quality depth data in a range of  environments is a challenging problem. 
In this paper, we innovate beyond existing approaches, replacing the use of explicit depth data during training with easier-to-obtain binocular stereo footage.

We propose a novel training objective that enables our convolutional neural network to learn to perform single image depth estimation, despite the absence of ground truth depth data.
Exploiting epipolar geometry constraints, we generate disparity images by training our network with an image reconstruction loss.
We show that solving for image reconstruction alone results in poor quality depth images. 
To overcome this problem, we propose a novel training loss that enforces consistency between the disparities produced relative to both the left and right images, leading to improved performance and robustness compared to existing approaches. 
Our method produces state of the art results for monocular depth estimation on the KITTI driving dataset, even outperforming supervised methods that have been trained with ground truth depth. 
\end{abstract}

%
%
\section{Introduction}\label{sec:introduction}
Depth estimation from images has a long history in computer vision. 
Fruitful approaches have relied on structure from motion, shape-from-X, binocular, and multi-view stereo. 
However, most of these techniques rely on the assumption that multiple observations of the scene of interest are available.
These can come in the form of multiple viewpoints, 
or observations of the scene under different lighting conditions. 
To overcome this limitation, there has recently been a surge in the number of works that pose the task of monocular depth estimation as a supervised learning problem \cite{ladicky2014pulling, eigen2014depth, liu2015learning}. 
These methods attempt to directly predict the depth of each pixel in an image using models that have been trained offline on large collections of ground truth depth data. 
While these methods have enjoyed great success, to date they have been restricted to scenes where large image collections and their corresponding pixel depths are available. 

Understanding the shape of a scene from a single image, independent of its appearance, is a fundamental problem in machine perception. 
There are many applications such as synthetic object insertion in computer graphics \cite{karsch2014automatic}, synthetic depth of field in computational photography \cite{Barron2015A}, grasping in robotics \cite{lenz2015deep}, using depth as a cue in human body pose estimation \cite{shotton2013real}, robot assisted surgery \cite{stoyanov2010real}, and automatic 2D to 3D conversion in film \cite{xie2016deep3d}.
Accurate depth data from one or more cameras is also crucial for self-driving cars, where expensive laser-based systems are often used.

\begin{figure}[t]
  \centering
  \input{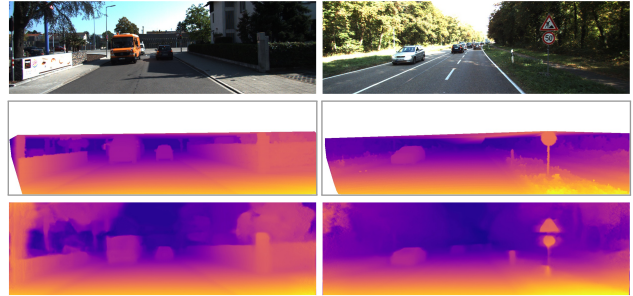}
  \vspace{5pt}
  \caption{Our depth prediction results on KITTI 2015. Top to bottom: input image, ground truth disparities, and our result. Our method is able to estimate depth for thin structures such as street signs and poles.}
  \label{fig:overview_results}
  \vspace{-10pt}
\end{figure}

Humans perform well at monocular depth estimation by exploiting cues such as perspective, scaling relative to the known size of familiar objects, appearance in the form of lighting and shading and occlusion \cite{howard2012perceiving}.
This combination of both top-down and bottom-up cues appears to link full scene understanding with our ability to accurately estimate depth.  
In this work, we take an alternative approach and treat automatic depth estimation as an image reconstruction problem during training.
Our fully convolutional model does not require any depth data, and is instead trained to synthesize depth as an intermediate. It learns to predict the pixel-level correspondence between pairs of rectified stereo images that have a known camera baseline.
There are some existing methods that also address the same problem, but with several limitations. For example they are not fully differentiable, making training suboptimal \cite{garg2016unsupervised}, or have image formation models that do not scale to large output resolutions \cite{xie2016deep3d}.
We improve upon these methods with a novel training objective and enhanced network architecture that significantly increases the quality of our final results. 
An example result from our algorithm is illustrated in Fig.~\ref{fig:overview_results}. 
Our method is fast and only takes on the order of $35$ milliseconds to predict a dense depth map for a $512\times 256$ image on a modern GPU.
Specifically, we propose the following contributions:
\newline1) A network architecture that performs end-to-end unsupervised monocular depth estimation with a novel training loss that enforces left-right depth consistency inside the network.
\newline2) An evaluation of several training losses and image formation models highlighting the effectiveness of our approach.
\newline3) In addition to showing state of the art results on a challenging driving dataset, we also show that our model generalizes to three different datasets, including a new outdoor urban dataset that we have collected ourselves, which we make openly available.
%
%
\section{Related Work}
There is a large body of work that focuses on depth estimation from images, either using pairs \cite{scharstein2002taxonomy}, several overlapping images captured from different viewpoints \cite{furukawa2015multi}, temporal sequences \cite{ranftldense}, or assuming a fixed camera, static scene, and changing lighting \cite{woodham1980photometric, abrams2012heliometric}.
These approaches are typically only applicable when there is more than one input image available of the scene of interest. 
Here we focus on works related to monocular depth estimation, where there is only a single input image, and no assumptions about the scene geometry or types of objects present are made. 

\subsection*{Learning-Based Stereo}
The vast majority of stereo estimation algorithms have a data term which computes the similarity between each pixel in the first image and every other pixel in the second image.
Typically the stereo pair is rectified and thus the problem of disparity (\ie scaled inverse depth) estimation can be posed as a 1D search problem for each pixel. 
Recently, it has been shown that instead of using hand defined similarity measures, treating the matching as a supervised learning problem and training a function to predict the correspondences produces far superior results \cite{vzbontar2016stereo, ladicky2015learning}.
It has also been shown that posing this binocular correspondence search as a multi-class classification problem has advantages both in terms of quality of results and speed \cite{luo16a}.
Instead of just learning the matching function, Mayer \ea\cite{mayer2015large} introduced a fully convolutional \cite{shelhamer2016fully} deep network called DispNet that directly computes the correspondence field between two images.
At training time, they attempt to directly predict the disparity for each pixel by minimizing a regression training loss. 
DispNet has a similar architecture to their previous end-to-end deep optical flow network \cite{fischer2015flownet}.

The above methods rely on having large amounts of accurate ground truth disparity data and stereo image pairs at training time.
This type of data can be difficult to obtain for real world scenes, so these approaches typically use synthetic data for training. 
Synthetic data is becoming more realistic, \eg\cite{gaidon2016virtual}, but still requires the manual creation of new content for every new application scenario.

\subsection*{Supervised Single Image Depth Estimation}
Single-view, or monocular, depth estimation refers to the problem setup where only a single image is available at test time. 
Saxena \ea\cite{saxena2009make3d} proposed a patch-based model known as Make3D that first over-segments the input image into patches and then estimates the 3D location and orientation of local planes to explain each patch.
The predictions of the plane parameters are made using a linear model trained offline on a dataset of laser scans, and the predictions are then combined together using an MRF. 
The disadvantage of this method, and other planar based approximations, \eg \cite{hoiem2005automatic}, is that they can have difficulty modeling thin structures and, as predictions are made locally, lack the global context required to generate realistic outputs.
Instead of hand-tuning the unary and pairwise terms, Liu \ea\cite{liu2015learning} use a  convolutional neural network (CNN) to learn them. 
In another local approach, Ladicky \ea\cite{ladicky2014pulling} incorporate semantics into their model to improve their per pixel depth estimation.
Karsch \ea\cite{karsch2014depth} attempt to produce more consistent image level predictions by copying whole depth images from a training set.
A drawback of this approach is that it requires the entire training set to be available at test time. 

Eigen \ea\cite{eigen2014depth, eigen2015predicting} showed that it was possible to produce dense pixel depth estimates using a two scale deep network trained on images and their corresponding depth values.
Unlike most other previous work in single image depth estimation, they do not rely on hand crafted features or an initial over-segmentation and instead learn a representation directly from the raw pixel values. 
Several works have built upon the success of this approach using techniques such as CRFs to improve accuracy \cite{li2015depth}, changing the loss from regression to classification \cite{cao2016estimating}, using other more robust loss functions \cite{laina2016deeper}, and incorporating strong scene priors in the case of the related problem of surface normal estimation \cite{wang2015designing}. 
Again, like the previous stereo methods, these approaches rely on having high quality, pixel aligned, ground truth depth at training time. 
We too perform single depth image estimation, but train with an added binocular color image, instead of requiring ground truth depth.

\subsection*{Unsupervised Depth Estimation}
Recently, a small number of deep network based methods for novel view synthesis and depth estimation have been proposed, which do not require ground truth depth at training time. 
Flynn \ea\cite{flynn2015deepstereo} introduced a novel image synthesis network called DeepStereo that generates new views by selecting pixels from nearby images. 
During training, the relative pose of multiple cameras is used to predict the appearance of a held-out nearby image. Then the most appropriate depths are selected to sample colors from the neighboring images, based on plane sweep volumes.
At test time, image synthesis is performed on small overlapping patches. 
As it requires several nearby posed images at test time DeepStereo is not suitable for monocular depth estimation. 

The Deep3D network of Xie \ea\cite{xie2016deep3d} also addresses the problem of novel view synthesis, where their goal is to generate the corresponding right view from an input left image (\ie the source image) in the context of binocular pairs.
Again using an image reconstruction loss, their method produces a distribution over all the possible disparities for each pixel. 
The resulting synthesized right image pixel values are a combination of the pixels on the same scan line from the left image, weighted by the probability of each disparity. 
The disadvantage of their image formation model is that increasing the number of candidate disparity values greatly increases the memory consumption of the algorithm, making it difficult to scale their approach to bigger output resolutions.
In this work, we perform a comparison to the Deep3D image formation model, and show that our algorithm produces superior results. 

Closest to our model in spirit is the concurrent work of Garg \ea\cite{garg2016unsupervised}.
Like Deep3D and our method, they train a network for monocular depth estimation using an image reconstruction loss. 
However, their image formation model is not fully differentiable. To compensate, they perform a Taylor approximation to linearize their loss resulting in an objective that is more challenging to optimize. 
Similar to other recent work, \eg\cite{patraucean2015spatio, zhou2016learning, zhou2016view}, our model overcomes this problem by using bilinear sampling \cite{jaderberg2015spatial} to generate images, resulting in a fully (sub-)differentiable training loss.

We propose a fully convolutional deep neural network loosely inspired by the supervised DispNet architecture of Mayer \ea\cite{mayer2015large}. 
By posing monocular depth estimation as an image reconstruction problem, we can solve for the disparity field without requiring ground truth depth. 
However, only minimizing a photometric loss can result in good quality image reconstructions but poor quality depth.
Among other terms, our fully differentiable training loss includes a left-right consistency check to improve the quality of our synthesized depth images. 
This type of consistency check is commonly used as a post-processing step in many stereo methods, \eg \cite{vzbontar2016stereo}, but we incorporate it directly into our network. 
%
%

\section{Method}
This section describes our single image depth prediction network.
We introduce a novel depth estimation training loss, featuring an inbuilt left-right consistency check, which enables us to train on image pairs without requiring supervision in the form of ground truth depth. 

\begin{figure}
  \centering
  \includegraphics[width=0.9\linewidth]{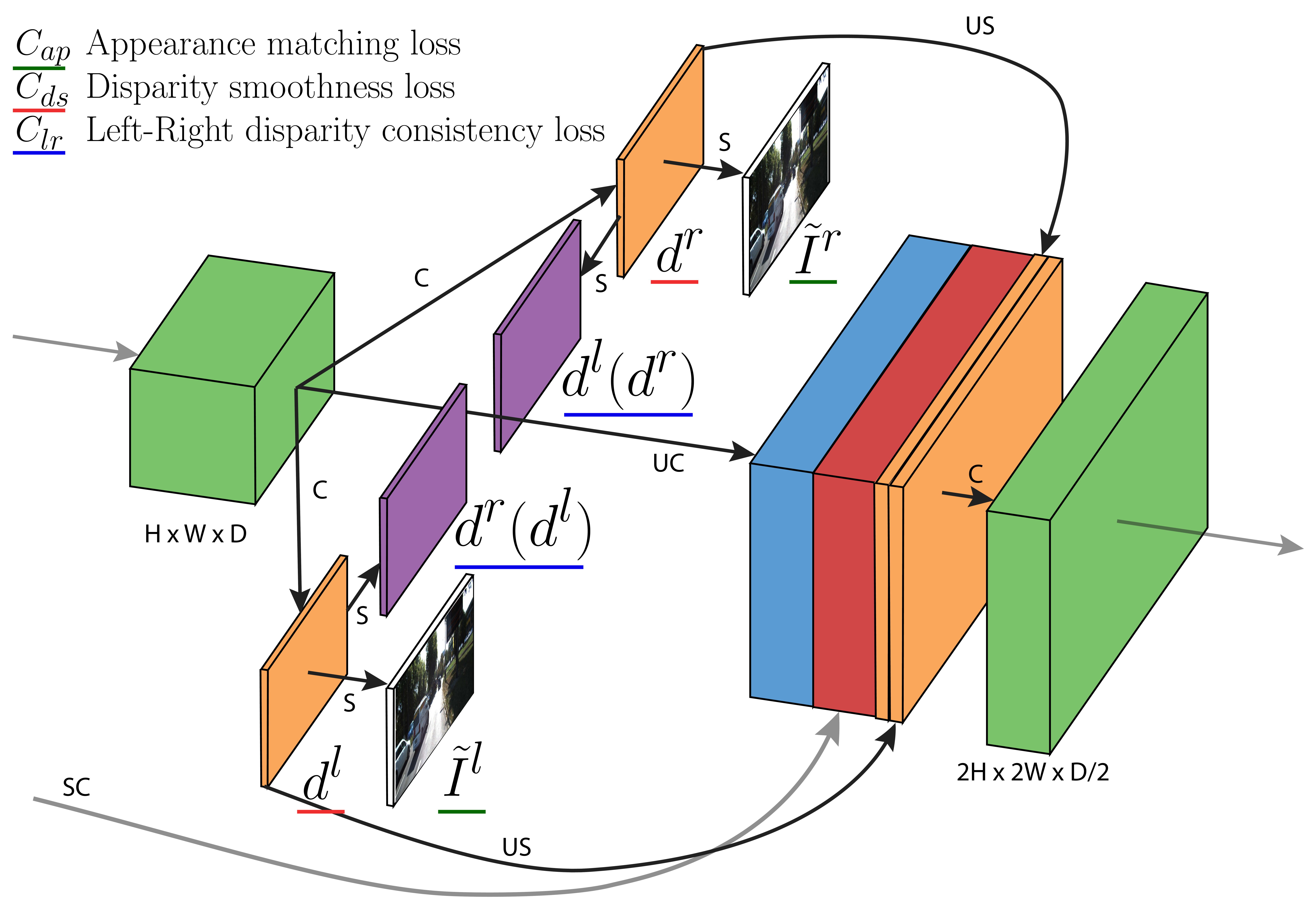}
  \caption{Our loss module outputs left and right disparity maps, $d^l$ and $d^r$. The loss combines smoothness, reconstruction, and left-right disparity consistency terms. This same module is repeated at each of the four different output scales. C: Convolution, UC: Up-Convolution, S: Bilinear Sampling, US: Up-Sampling, SC: Skip Connection. \vspace{-5pt}}
  \label{fig:pipeline}
\end{figure}

\subsection{Depth Estimation as Image Reconstruction}
Given a single image $I$ at test time, our goal is to learn a function $f$ that can predict the per-pixel scene depth, $\hat{d} = f(I)$. 
Most existing learning based approaches treat this as a supervised learning problem, where they have color input images and their corresponding target depth values at training.
It is presently not practical to acquire such ground truth depth data for a large variety of scenes. Even expensive hardware, such as laser scanners, can be imprecise in natural scenes featuring movement and reflections.
As an alternative, we instead pose depth estimation as an image reconstruction problem during training.
The intuition here is that, given a calibrated pair of binocular cameras, if we can learn a function that is able to reconstruct one image from the other, then we have learned something about the $3$D shape of the scene that is being imaged.

Specifically, at training time, we have access to two images $I^l$ and $I^r$, corresponding to the left and right color images from a calibrated stereo pair, captured at the same moment in time. 
Instead of trying to directly predict the depth, we attempt to find the dense correspondence field $d^r$ that, when applied to the left image, would enable us to reconstruct the right image. 
We will refer to the reconstructed image $I^l(d^r)$ as $\tilde{I}^r$. 
Similarly, we can also estimate the left image given the right one, $\tilde{I}^l = I^r(d^l)$.
Assuming that the images are rectified \cite{hartley2003multiple}, $d$ corresponds to the image disparity - a scalar value per pixel that our model will learn to predict.   
Given the baseline distance $b$ between the cameras and the camera focal length $f$,  we can then trivially recover the depth $\hat{d}$ from the predicted disparity, $\hat{d} = bf/d$. 

\subsection{Depth Estimation Network}

\begin{figure}[!t]
  \centering
  \includegraphics[width=0.85\linewidth]{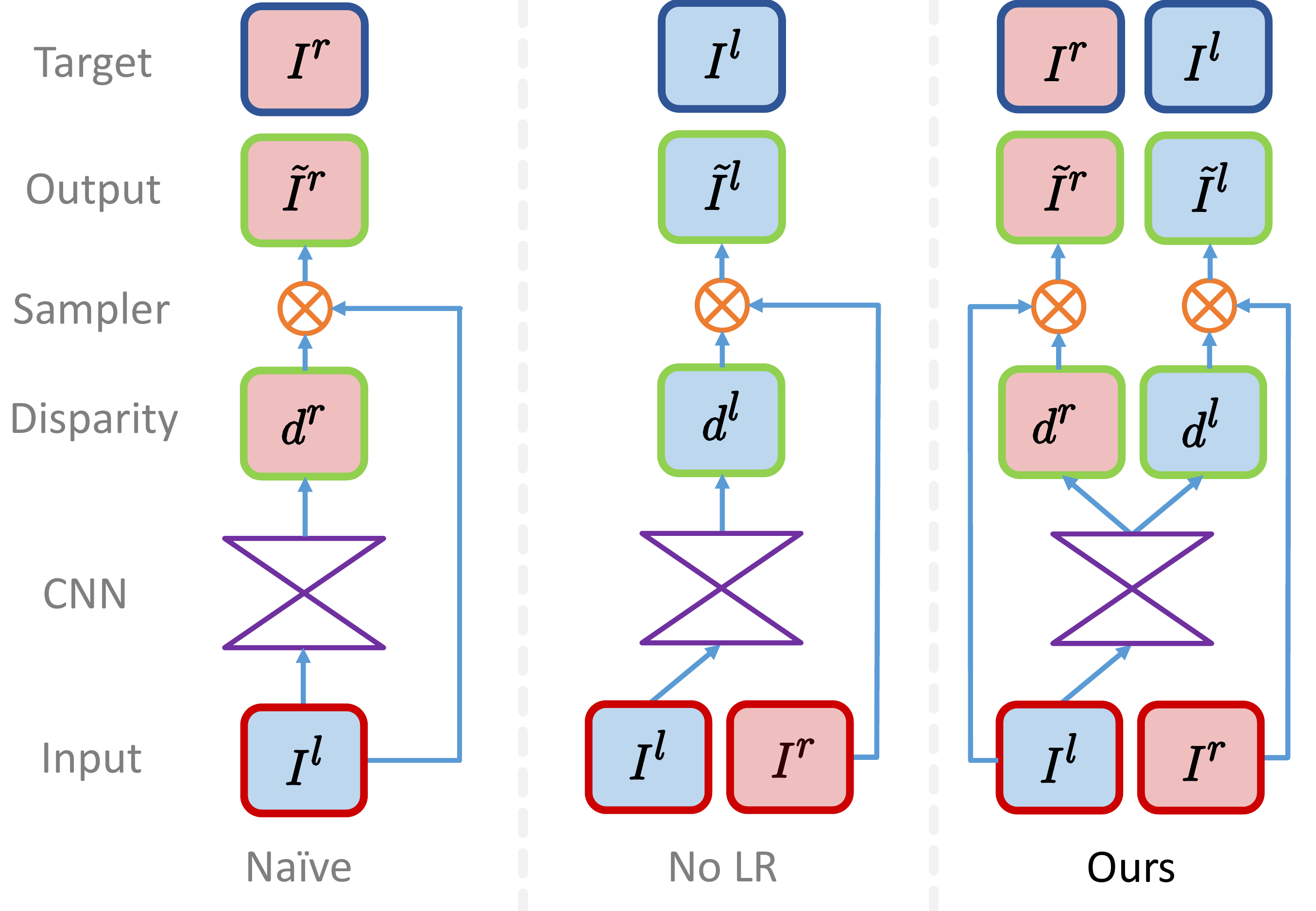}
  \caption{Sampling strategies for backward mapping. With na\"{i}ve sampling the CNN produces a disparity map aligned with the target instead of the input. No LR corrects for this, but suffers from artifacts. Our approach uses the left image to produce disparities for both images, improving quality by enforcing mutual consistency.}
  \label{fig:LR}
\end{figure}

At a high level, our network estimates depth by inferring the disparities that warp the left image to match the right one. 
The key insight of our method is that we can simultaneously infer both disparities (left-to-right and right-to-left), using only the left input image, and obtain better depths by enforcing them to be consistent with each other.

Our network generates the predicted image with backward mapping using a bilinear sampler, resulting in a fully differentiable image formation model. 
As illustrated in Fig.~\ref{fig:LR}, na\"ively learning to generate the right image by sampling from the left one will produce disparities aligned with the right image (target).
However, we want the output disparity map to align with the input left image, meaning the network has to sample from the right image.
We could instead train the network to generate the left view by sampling from the right image, thus creating a left view aligned disparity map (\textbf{No LR} in Fig.~\ref{fig:LR}). 
While this alone works, the inferred disparities exhibit `texture-copy' artifacts and errors at depth discontinuities as seen in Fig. \ref{fig:lr_consistency_results}.
We solve this by training the network to predict the disparity maps for both views by sampling from the opposite input images. 
This still only requires a single left image as input to the convolutional layers and the right image is only used during training (\textbf{Ours} in Fig.~\ref{fig:LR}). 
Enforcing consistency between both disparity maps using this novel left-right consistency cost leads to more accurate results.

Our fully convolutional architecture is inspired by DispNet~\cite{mayer2015large}, but features several important modifications that enable us to train without requiring ground truth depth. 
Our network, is composed of two main parts - an encoder (from cnv1 to cnv7b) and decoder (from upcnv7), please see the supplementary material for a detailed description.
The decoder uses skip connections \cite{shelhamer2016fully} from the encoder's activation blocks, enabling it to resolve higher resolution details. 
We output disparity predictions at four different scales (disp4 to disp1), which double in spatial resolution at each of the subsequent scales. 
Even though it only takes a single image as input, our network predicts two disparity maps at each output scale - left-to-right and right-to-left.

\begin{table*}[!h]
  \centering
  \resizebox{0.94\textwidth}{!}{
  \begin{tabular}{|l|c||c|c|c|c|c|c|c|c|}
  \hline
  Method & Dataset & \cellcolor{col1}Abs Rel & \cellcolor{col1}Sq Rel &  \cellcolor{col1}RMSE  & \cellcolor{col1}RMSE log & \cellcolor{col1}{\it D1-all} & \cellcolor{col2}$\delta < 1.25 $ & \cellcolor{col2}$\delta < 1.25^{2}$ & \cellcolor{col2}$\delta < 1.25^{3}$\\
  \hline 
  Ours with Deep3D \cite{xie2016deep3d} & K & 0.412 & 16.37 & 13.693 & 0.512 & 66.85 & 0.690 & 0.833 & 0.891\\
  Ours with Deep3Ds \cite{xie2016deep3d} & K & 0.151 & 1.312 & 6.344 & 0.239 & 59.64 & 0.781 & 0.931 & 0.976\\ 
  Ours No LR & K & 0.123 & 1.417 & 6.315 & 0.220 & 30.318 & 0.841 & 0.937 & 0.973 \\ 
  Ours & K  & 0.124 & 1.388 & 6.125 & 0.217 & 30.272 & 0.841 & 0.936 & 0.975 \\ 
  Ours & CS & 0.699 & 10.060 & 14.445 & 0.542 & 94.757 & 0.053 & 0.326 & 0.862\\ 
  Ours & CS + K & 0.104 & 1.070 & 5.417 & 0.188 & 25.523 & 0.875 & 0.956 & 0.983\\
  Ours pp & CS + K & 0.100 & 0.934 & 5.141 & 0.178 & 25.077 & 0.878 & 0.961 & \textbf{0.986}\\
  Ours resnet pp & CS + K & \textbf{0.097} & \textbf{0.896} & \textbf{5.093} & \textbf{0.176} & \textbf{23.811} & \textbf{0.879} & \textbf{0.962} & \textbf{0.986}\\
  \hline
  Ours Stereo & K  & 0.068 & 0.835 & 4.392 & 0.146 & 9.194 & 0.942 & 0.978 & 0.989\\
  \hline
  \end{tabular}	  
  \begin{tabular}{|l|}
    \hline
      \cellcolor{col1} Lower is better\\ \hline
      \\ \hline
      \cellcolor{col2} Higher is better\\ \hline   
  \end{tabular}
  }
  \vspace{10pt}
    \caption{Comparison of different image formation models. 
  Results on the KITTI 2015 stereo 200 training set disparity images \cite{Geiger2012CVPR}. 
  For training, K is the KITTI dataset \cite{Geiger2012CVPR} and CS is Cityscapes \cite{Cordts2016Cityscapes}.
  Our model with left-right consistency performs the best, and is further improved with the addition of the Cityscapes data. 
  The last row shows the result of our model trained \emph{and tested} with two input images instead of one (see Sec.~\ref{sec:Stereo}).}
    \label{tab:kitti_official}
    \vspace{-10pt}
\end{table*}

\subsection{Training Loss}
We define a loss $C_s$ at each output scale s, forming the total loss as the sum $C = \sum_{s=1}^4 C_s$.
Our loss module (Fig.~\ref{fig:pipeline}) computes $C_s$ as a combination of three main terms, 
\begin{equation}
C_s = \alpha_{ap} (C_{ap}^l + C_{ap}^r) + \alpha_{ds} (C_{ds}^l + C_{ds}^r) + \alpha_{lr} (C_{lr}^l + C_{lr}^r),
\label{eq:cs}
\end{equation}
where $C_{ap}$ encourages the reconstructed image to appear similar to the corresponding training input, $C_{ds}$ enforces smooth disparities, and $C_{lr}$ prefers the predicted left and right disparities to be consistent. Each of the main terms contains both a left and a right image variant, but only the left image is fed through the convolutional layers. 

Next, we present each component of our loss in terms of the left image (\eg $C_{ap}^l$). The right image versions, \eg $C_{ap}^r$, require to swap left for right and to sample in the opposite direction.

\paragraph*{Appearance Matching Loss}
During training, the network learns to generate an image by sampling pixels from the opposite stereo image. 
Our image formation model uses the image sampler from the spatial transformer network (STN) \cite{jaderberg2015spatial} to sample the input image using a disparity map. 
The STN uses bilinear sampling where the output pixel is the weighted sum of four input pixels. 
In contrast to alternative approaches \cite{garg2016unsupervised,xie2016deep3d}, the bilinear sampler used is locally fully differentiable and integrates seamlessly into our fully convolutional architecture.
This means that we do not require any simplification or approximation of our cost function.

Inspired by \cite{lossfunctions}, we use a combination of an $L1$ and single scale SSIM \cite{wang2004image} term as our photometric image reconstruction cost $C_{ap}$, which compares the input image $I^l_{ij}$ and its reconstruction $\tilde{I}^l_{ij}$, where $N$ is the number of pixels,
\begin{equation}C_{ap}^l = \frac{1}{N} \sum_{i,j} \alpha \frac{1 - \textup{SSIM}(I^l_{ij}, \tilde{I}^l_{ij})}{2} + (1-\alpha)\left \| I^l_{ij} - \tilde{I}^l_{ij} \right \|.
\label{eq:ca}
\end{equation}
Here, we use a simplified SSIM with a $3\times3$ block filter instead of a Gaussian, and set $\alpha = 0.85$.

\paragraph*{Disparity Smoothness Loss}
We encourage disparities to be locally smooth with an $L1$ penalty on the disparity gradients $\partial d$. 
As depth discontinuities often occur at image gradients, similar to \cite{heise2013pm}, we weight this cost with an edge-aware term using the image gradients $\partial I$,
\vspace{-3pt}
\begin{equation}C_{ds}^l = \frac{1}{N} \sum_{i,j} \left | \partial_x d^l_{ij}   \right | e^{-\left \| \partial_x I_{ij}^l \right \|} + \left | \partial_y d^l_{ij}   \right | e^{-\left \| \partial_y I^l_{ij} \right \|}.
\label{eq:cds}
\end{equation}

\paragraph*{Left-Right Disparity Consistency Loss}
To produce more accurate disparity maps, we train our network to predict both the left and right image disparities, while only being given the left view as input to the convolutional part of the network. 
To ensure coherence, we introduce an $L1$ left-right disparity consistency penalty as part of our model. 
This cost attempts to make the left-view disparity map be equal to the \emph{projected} right-view disparity map,
\begin{equation}C_{lr}^l = \frac{1}{N} \sum_{i,j} \left | d^l_{ij} -  d^r_{ij+d^l_{ij}} \right |.
\label{eq:clr}
\end{equation}
Like all the other terms, this cost is mirrored for the right-view disparity map and is evaluated at all of the output scales.

At test time, our network predicts the disparity at the finest scale level for the left image $d^l$, which has the same resolution as the input image. 
Using the known camera baseline and focal length from the training set, we then convert from the disparity map to a depth map.  
While we also estimate the right disparity $d^r$ during training, it is not used at test time. 

\begin{figure*}[!h]
  \centering
  \resizebox{\textwidth}{!}{
  \newcommand{\turnheightnew}{0.195\columnwidth}

\centering

\begin{tabular}{@{\hskip 1mm}c@{\hskip 1mm}c@{\hskip 1mm}c@{\hskip 1mm}c@{\hskip 1mm}c@{\hskip 1mm}c@{}}

\Large{Input} & \Large{GT} & \Large{Eigen \ea\cite{eigen2014depth}} & \Large{Liu \ea\cite{liu2015learning}} & \Large{Garg \ea\cite{garg2016unsupervised}} & \Large{Ours} \\

\includegraphics[height=\turnheightnew]{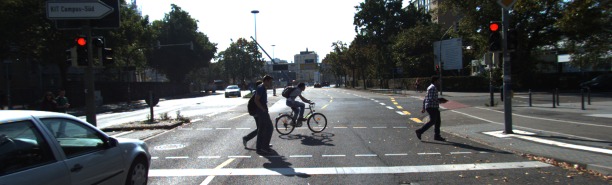} &
\fcolorbox{grey}{white}{\includegraphics[height=\turnheightnew]{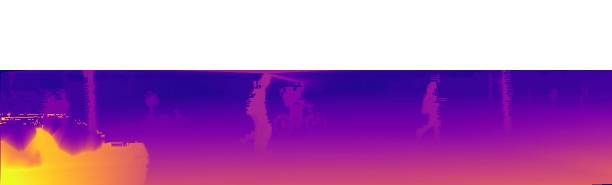}} &
\fcolorbox{grey}{white}{\includegraphics[height=\turnheightnew]{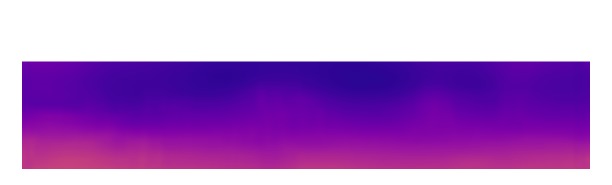}} &
\includegraphics[height=\turnheightnew]{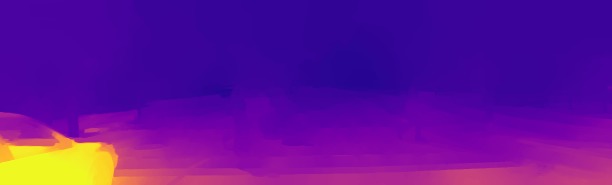} &
\includegraphics[height=\turnheightnew]{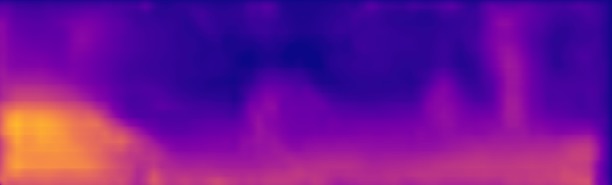} &
\includegraphics[height=\turnheightnew]{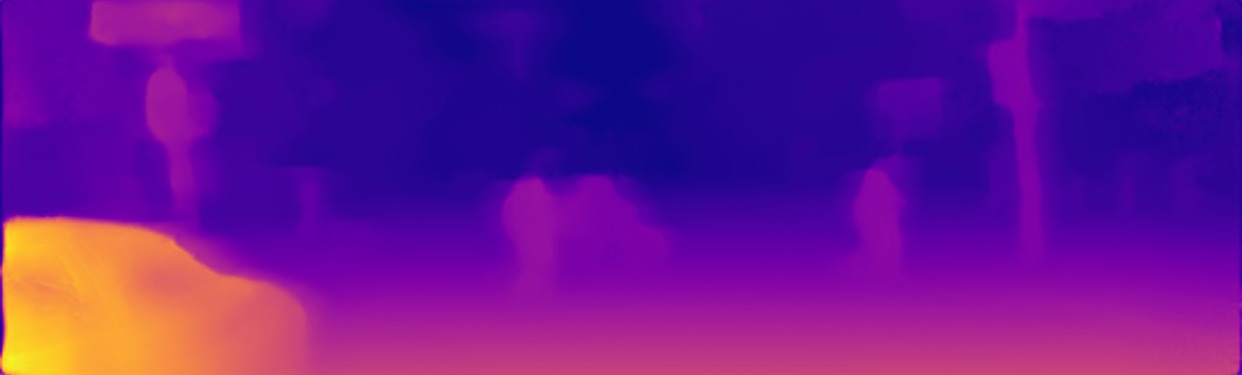} \\

\includegraphics[height=\turnheightnew]{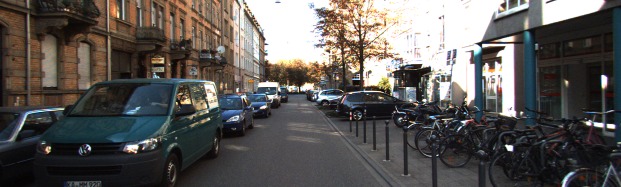} &
\fcolorbox{grey}{white}{\includegraphics[height=\turnheightnew]{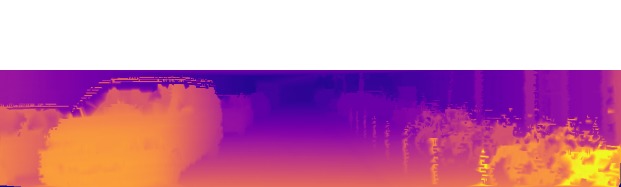}} &
\fcolorbox{grey}{white}{\includegraphics[height=\turnheightnew]{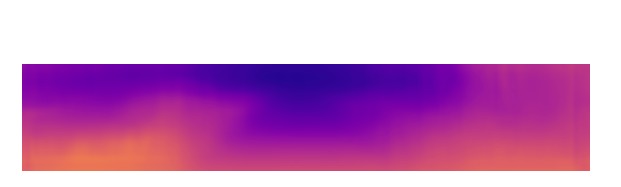}} &
\includegraphics[height=\turnheightnew]{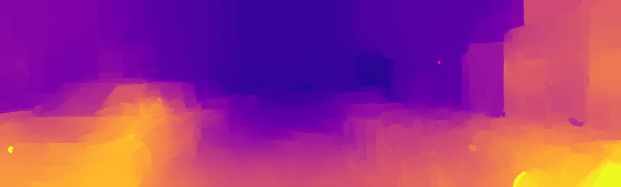} &
\includegraphics[height=\turnheightnew]{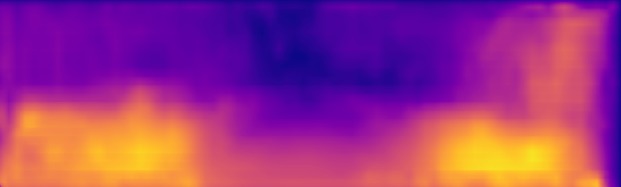} &
\includegraphics[height=\turnheightnew]{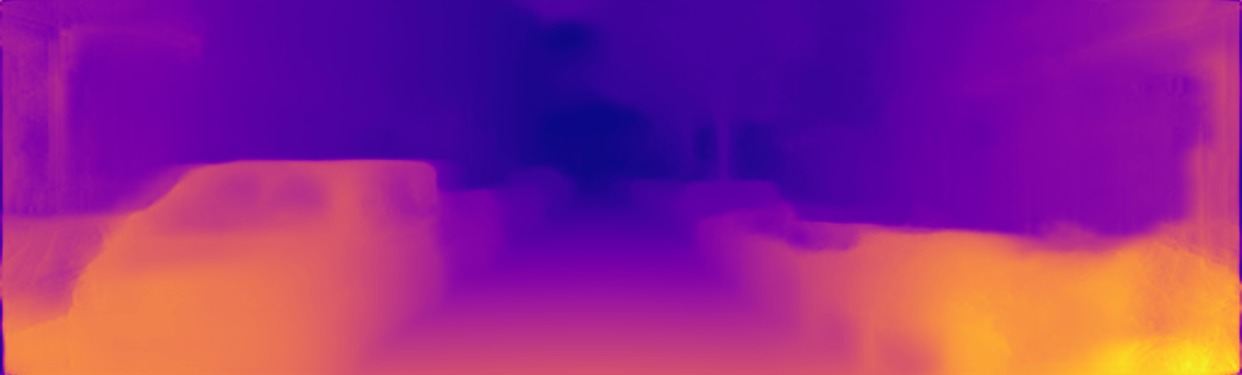} \\ 

\includegraphics[height=\turnheightnew]{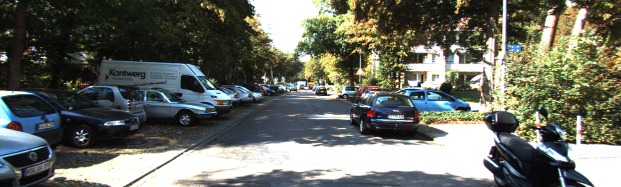} &
\fcolorbox{grey}{white}{\includegraphics[height=\turnheightnew]{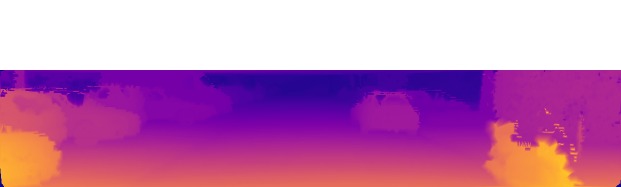}} &
\fcolorbox{grey}{white}{\includegraphics[height=\turnheightnew]{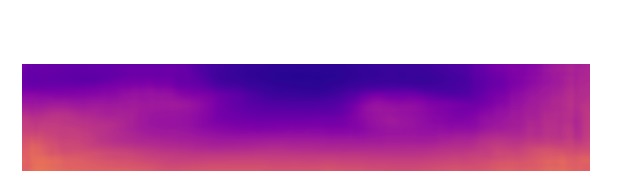}} &
\includegraphics[height=\turnheightnew]{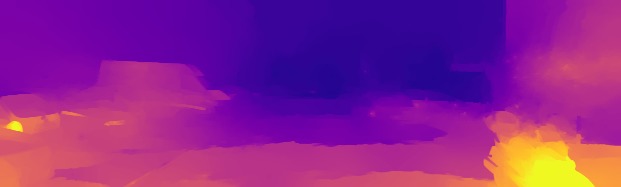} &
\includegraphics[height=\turnheightnew]{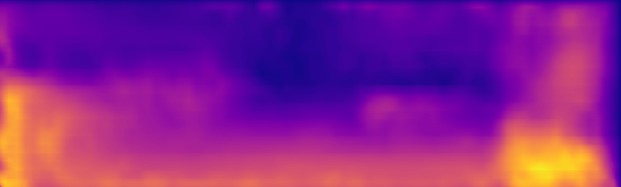} &
\includegraphics[height=\turnheightnew]{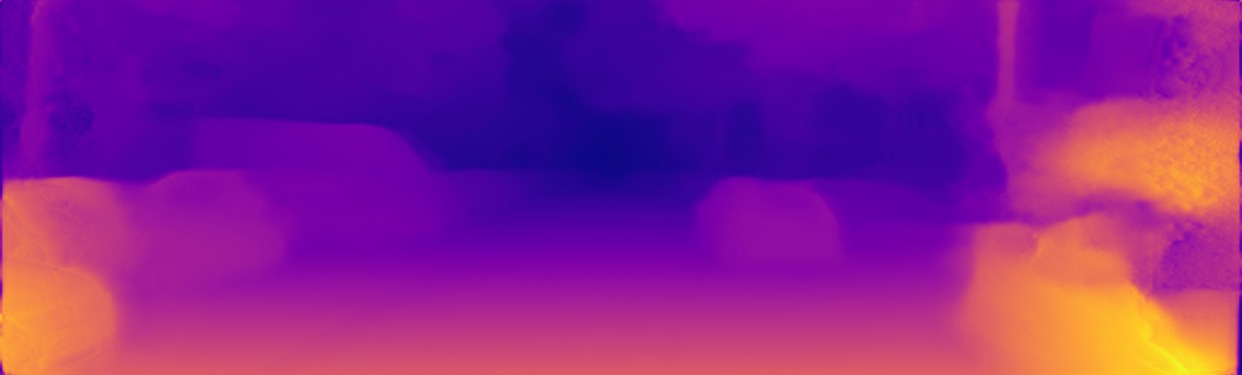} \\ 

\end{tabular}







 }
  \vspace{0pt}
  \caption{Qualitative results on the KITTI Eigen Split. The ground truth velodyne depth being very sparse,  we interpolate it for visualization purposes. Our method does better at resolving small objects such as the pedestrians and poles.}
  \label{fig:kitti_eigen}    
\end{figure*}

%
%
\section{Results}
Here we compare the performance of our approach to both supervised and unsupervised single view depth estimation methods. 
We train on rectified stereo image pairs, and do not require any supervision in the form of ground truth depth.
Existing single image datasets, such as \cite{Silberman:ECCV12, saxena2009make3d}, that lack stereo pairs, are not suitable for evaluation.
Instead we evaluate our approach using the popular KITTI 2015 \cite{Geiger2012CVPR} dataset.
To evaluate our image formation model, we compare to a variant of our algorithm that uses the original Deep3D~\cite{xie2016deep3d} image formation model and a modified one, Deep3Ds, with an added smoothness constraint. 
We also evaluate our approach with and without the left-right consistency constraint.

\subsection{Implementation Details}
The network which is implemented in TensorFlow \cite{tensorflow} contains $31$ million trainable parameters, and takes on the order of $25$ hours to train using a single Titan X GPU on a dataset of $30$ thousand images for $50$ epochs. 
Inference is fast and takes less than $35$ ms, or more than $28$ frames per second, for a $512\times 256$ image, including transfer times to and from the GPU. Please see the supplementary material and our code\footnote{Available at \url{https://github.com/mrharicot/monodepth}} for more details.

During optimization, we set the weighting of the different loss components to $\alpha_{ap} = 1$ and $\alpha_{lr} = 1$. 
The possible output disparities are constrained to be between $0$ and $d_{max}$ using a scaled sigmoid non-linearity, where $d_{max} = 0.3 \times$ the image width at a given output scale. 
As a result of our multi-scale output, the typical disparity of neighboring pixels will differ by a factor of two between each scale (as we are upsampling the output by a factor of two). 
To correct for this, we scale the disparity smoothness term $\alpha_{ds}$ with $r$ for each scale to get equivalent smoothing at each level. 
Thus $\alpha_{ds} = 0.1 / r $, where $r$ is the downscaling factor of the corresponding layer with respect to the resolution of the input image that is passed into the network.

For the non-linearities in the network, we used exponential linear units \cite{elus} instead of the commonly used rectified liner units (ReLU) \cite{nair2010rectified}. 
We found that ReLUs tended to prematurely fix the predicted disparities at intermediate scales to a single value, making subsequent improvement difficult. Following \cite{odena2016deconvolution}, we replaced the usual deconvolutions with a nearest neighbor upsampling followed by a convolutions.
We trained our model from scratch for $50$ epochs, with a batch size of $8$ using Adam \cite{adamsolver}, where $\beta_1 = 0.9$, $\beta_2 = 0.999$, and $\epsilon = 10^{-8}$. 
We used an initial learning rate of $\lambda = 10^{-4}$ which we kept constant for the first $30$ epochs before halving it every $10$ epochs until the end.
We initially experimented with progressive update schedules, as in \cite{mayer2015large}, where lower resolution image scales were optimized first.
However, we found that optimizing all four scales at once led to more stable convergence.
Similarly, we use an identical weighting for the loss of each scale as we found that weighting them differently led to unstable convergence.
We experimented with batch normalization \cite{ioffe2015batch}, but found that it did not produce a significant improvement, and ultimately excluded it.

Data augmentation is performed on the fly. We flip the input images horizontally with a $50\%$ chance, taking care to also swap both images so they are in the correct position relative to each other.
We also added color augmentations, with a $50\%$ chance, where we performed random gamma, brightness, and color shifts by sampling from uniform distributions in the ranges $[0.8, 1.2]$ for gamma, $[0.5, 2.0]$ for brightness, and $[0.8, 1.2]$ for each color channel separately.

\paragraph{Resnet50} For the sake of completeness, and similar to \cite{laina2016deeper}, we also show a variant of our model using Resnet50 \cite{he2016deep} as the encoder, the rest of the architecture, parameters and training procedure staying identical. This variant contains $48$ million trainable parameters and is indicated by \textbf{resnet} in result tables.

\paragraph{Post-processing} In order to reduce the effect of stereo disocclusions which create disparity ramps on both the left side of the image and of the occluders, a final post-processing step is performed on the output. 
For an input image $I$ at test time, we also compute the disparity map $d'_l$ for its horizontally flipped image $I'$. 
By flipping back this disparity map we obtain a disparity map $d''_l$, which aligns with $d_l$ but where the disparity ramps are located on the right of occluders as well as on the right side of the image. 
We combine both disparity maps to form the final result by assigning the first $5\%$ on the left of the image using $d''_l$ and the last $5\%$ on the right to the disparities from $d_l$. 
The central part of the final disparity map is the average of $d_l$ and $d'_l$. 
This final post-processing step leads to both better accuracy and less visual artifacts at the expense of doubling the amount of test time computation. We indicate such results using \textbf{pp} in result tables.

\begin{table*}[t]
  \centering
  \resizebox{0.94\textwidth}{!}{
  \begin{tabular}{|l|c|c||c|c|c|c|c|c|c|}
  \hline
  Method & Supervised & Dataset & \cellcolor{col1}Abs Rel & \cellcolor{col1}Sq Rel & \cellcolor{col1}RMSE  & \cellcolor{col1}RMSE log & \cellcolor{col2}$\delta < 1.25 $ & \cellcolor{col2}$\delta < 1.25^{2}$ & \cellcolor{col2}$\delta < 1.25^{3}$\\
  \hline 
  Train set mean & No & K & 0.361 & 4.826 & 8.102 & 0.377 & 0.638 & 0.804 & 0.894\\    
  Eigen \ea\cite{eigen2014depth} Coarse $^{\circ}$ & Yes & K & 0.214 & 1.605 & 6.563 & 0.292 & 0.673 & 0.884 & 0.957\\ 
  Eigen \ea\cite{eigen2014depth} Fine $^{\circ}$ & Yes & K & 0.203 & 1.548 & 6.307 & 0.282 & 0.702 & 0.890 & 0.958\\ 
  Liu \ea\cite{liu2015learning} DCNF-FCSP FT \mbox{*} & Yes & K & 0.201 & 1.584 & 6.471 & 0.273 & 0.68 & 0.898 & 0.967\\
  \textbf{Ours No LR} & No & K & 0.152 & 1.528 & 6.098 & 0.252 & 0.801 & 0.922 & 0.963\\ 
  \textbf{Ours} & No & K & 0.148 & 1.344 & 5.927 & 0.247 & 0.803 & 0.922 & 0.964\\ 
  \textbf{Ours} & No & CS + K & 0.124 & 1.076 & 5.311 & 0.219 & 0.847 & 0.942 & 0.973\\
   \textbf{Ours pp} & No & CS + K & 0.118 & 0.923 & 5.015 & 0.210 & 0.854 & 0.947 & \textbf{0.976}\\
   \textbf{Ours resnet pp} & No & CS + K & \textbf{0.114} & \textbf{0.898} & \textbf{4.935} & \textbf{0.206} & \textbf{0.861} & \textbf{0.949} & \textbf{0.976}\\
  \hline
  Garg \ea\cite{garg2016unsupervised} L12 Aug 8$\times$ cap 50m & No & K & 0.169 & 1.080 & 5.104 & 0.273 & 0.740 & 0.904 & 0.962 \\ 
  \textbf{Ours} cap 50m & No & K & 0.140 & 0.976 & 4.471 & 0.232 & 0.818 & 0.931 & 0.969\\ 
  \textbf{Ours} cap 50m & No & CS + K & 0.117 & 0.762 & 3.972 & 0.206 & 0.860 & 0.948 & 0.976\\
  \textbf{Ours pp} cap 50m & No & CS + K & 0.112 & 0.680 & 3.810 & 0.198 & 0.866 & 0.953 & \textbf{0.979}\\
  \textbf{Ours resnet pp} cap 50m & No & CS + K & \textbf{0.108} & \textbf{0.657} & \textbf{3.729} & \textbf{0.194} & \textbf{0.873} & \textbf{0.954} & \textbf{0.979}\\ 
  \hline
  \textbf{Our pp} uncropped & No & CS + K & 0.134 & 1.261 & 5.336 & 0.230 & 0.835 & 0.938 & 0.971\\
  \textbf{Ours resnet pp} uncropped & No & CS + K & 0.130 & 1.197 & 5.222 & 0.226 & 0.843 & 0.940 & 0.971\\
  \hline
  \end{tabular}
  
  \begin{tabular}{|l|}
    \hline
      \cellcolor{col1} Lower is better\\ \hline
      \\ \hline
      \cellcolor{col2} Higher is better\\ \hline   
  \end{tabular}
  
  }
  \vspace{10pt}
  \caption{Results on KITTI 2015 \cite{Geiger2012CVPR} using the split of Eigen \ea\cite{eigen2014depth}. 
  For training, K is the KITTI dataset \cite{Geiger2012CVPR} and CS is Cityscapes \cite{Cordts2016Cityscapes}.
  The predictions of Liu \ea\cite{liu2015learning}\mbox{*} are generated on a mix of the left and right images instead of just the left input images. 
  For a fair comparison, we compute their results relative to the correct image.
  As in the provided source code, Eigen \ea\cite{eigen2014depth}$^{\circ}$ results are computed relative to the velodyne instead of the camera. 
  Garg \ea\cite{garg2016unsupervised} results are taken directly from their paper. All results, except \cite{eigen2014depth}, use the crop from \cite{garg2016unsupervised}.
  We also show our results with the same crop and maximum evaluation distance. The last two rows are computed on the uncropped ground truth.}
    \label{tab:kitti_eigen}
    \vspace{-10pt}
\end{table*}

\subsection{KITTI}
We present results for the KITTI dataset \cite{Geiger2012CVPR} using two different test splits, to enable comparison to existing works. 
In its raw form, the dataset contains $42,382$ rectified stereo pairs from $61$ scenes, with a typical image being $1242\times375$ pixels in size.

\paragraph{KITTI Split}
First we compare different variants of our method including different image formation models and different training sets. 
We evaluate on the $200$ high quality disparity images provided as part of the official KITTI training set, which covers a total of $28$ scenes. The remaining $33$ scenes contain $30,159$ images from which we keep $29,000$ for training and the rest for evaluation.
While these disparity images are much better quality than the reprojected velodyne laser depth values, they have CAD models inserted in place of moving cars. 
These CAD models result in ambiguous disparity values on transparent surfaces such as car windows, and issues at object boundaries where the CAD models do not perfectly align with the images.
In addition, the maximum depth present in the KITTI dataset is on the order of $80$ meters, and we cap the maximum predictions of all networks to this value.
Results are computed using the depth metrics from \cite{eigen2014depth} along with the {\it D1-all} disparity error from KITTI \cite{Geiger2012CVPR}.
The metrics from \cite{eigen2014depth} measure error in both meters from the ground truth and the percentage of depths that are within some threshold from the correct value.
It is important to note that measuring the error in depth space while the ground truth is given in disparities leads to precision issues. 
In particular, the non-thresholded measures can be sensitive to the large errors in depth caused by prediction errors at small disparity values.

\begin{figure}[b]
  \centering
  \includegraphics[width=\linewidth]{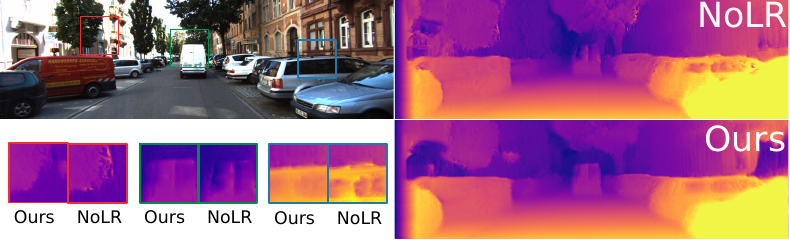}
  \caption{Comparison between our method with and without the left-right consistency. Our consistency term produces superior results on the object boundaries. Both results are shown without post-processing.}
  \label{fig:lr_consistency_results}     
\end{figure}

\begin{figure}
  \centering
  \centering
{\rotatebox{90}{\hspace{8pt}\footnotesize Input}}
\includegraphics[width=0.45\linewidth]{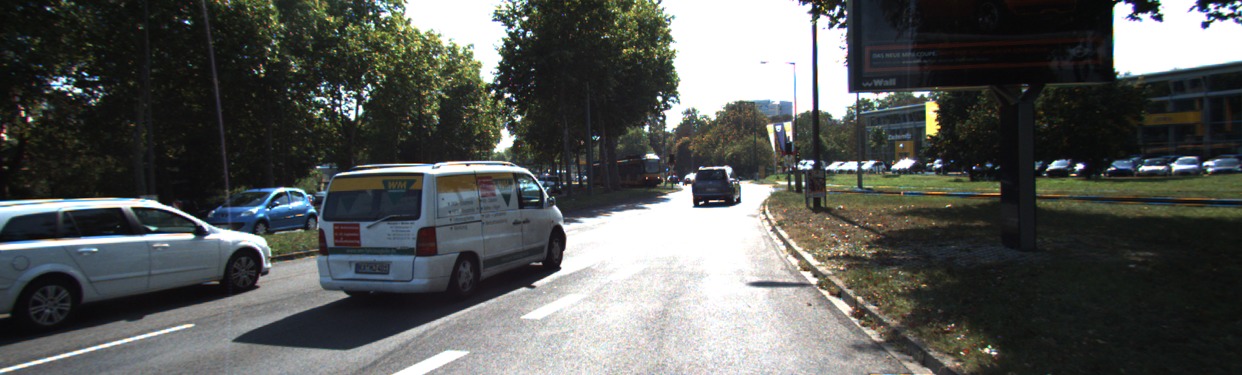} \\
\vspace{1mm}
\begin{tabular}{@{\hskip 1mm}c@{\hskip 1mm}c@{\hskip 1mm}c@{}}
{\rotatebox{90}{\hspace{2pt}\footnotesize Deep3D}} &
\includegraphics[width=0.45\linewidth]{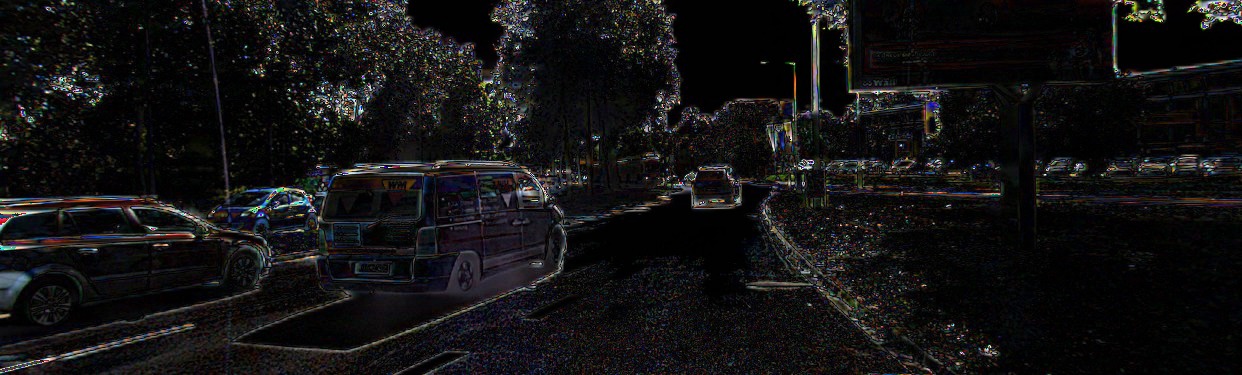} &
\includegraphics[width=0.45\linewidth]{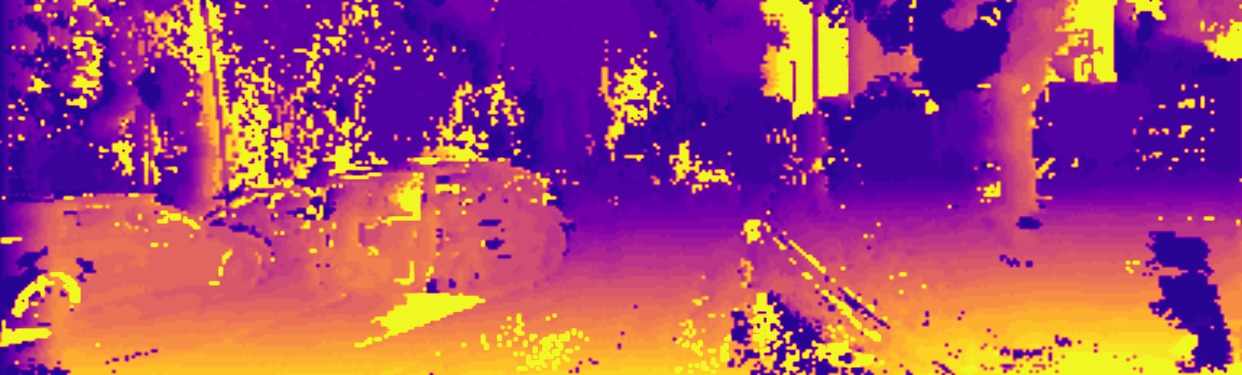} \\ 

{\rotatebox{90}{\hspace{1pt}\footnotesize Deep3Ds}} &
\includegraphics[width=0.45\linewidth]{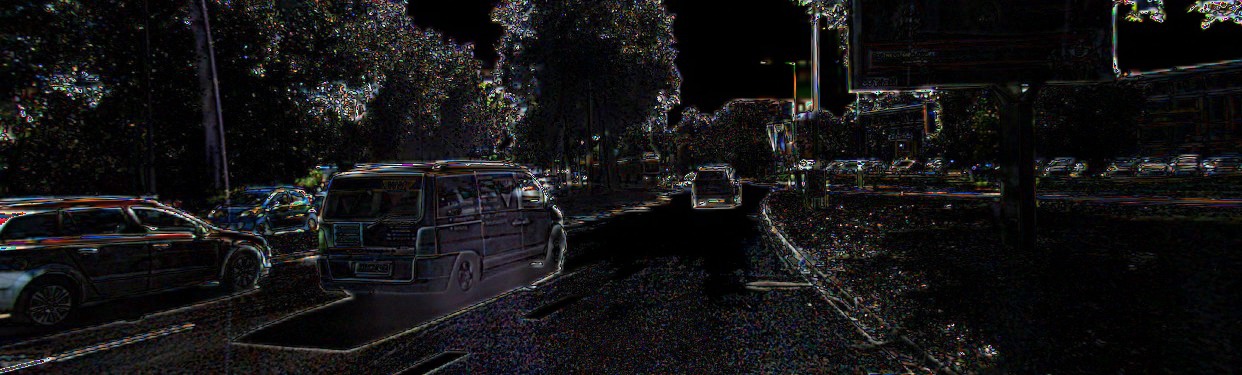} &
\includegraphics[width=0.45\linewidth]{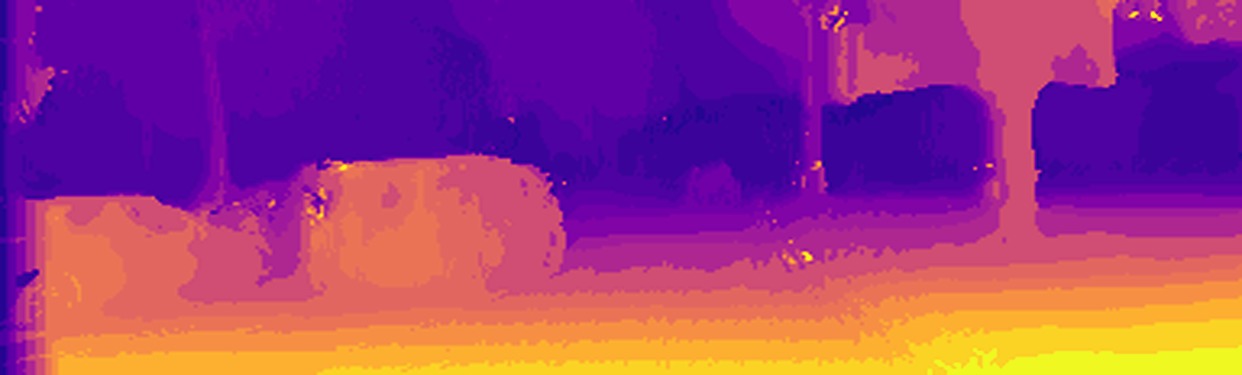} \\

{\rotatebox{90}{\hspace{8pt}\footnotesize Ours}} &
\includegraphics[width=0.45\linewidth]{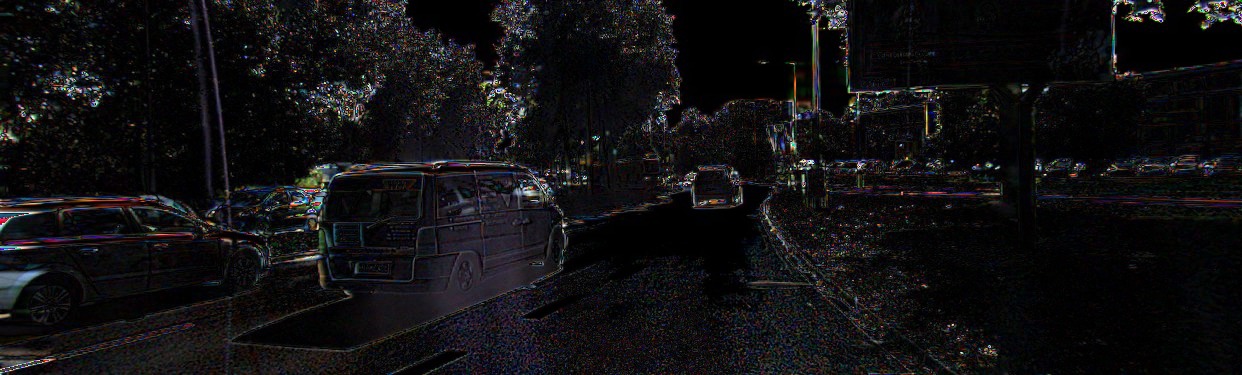} &
\includegraphics[width=0.45\linewidth]{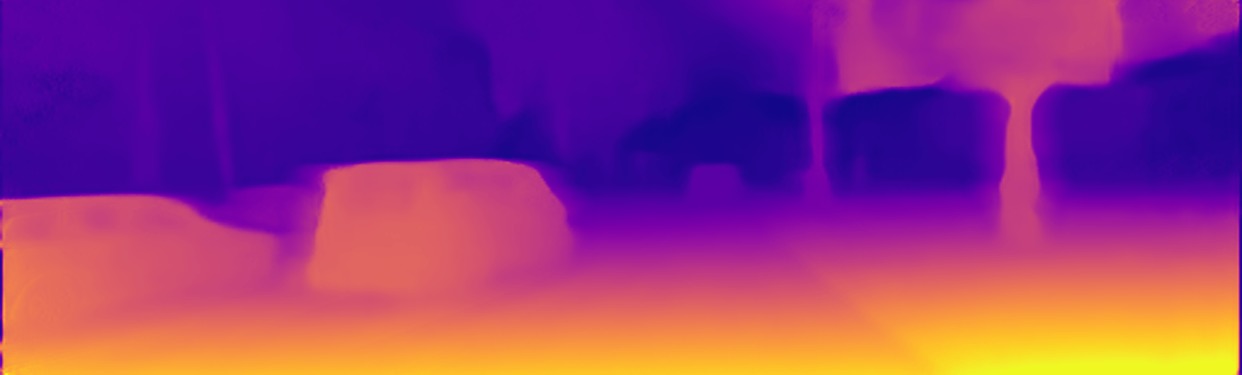} \\

 & Reconstruction error (x2) & Disparities \\
\end{tabular}
  \caption{Image reconstruction error on KITTI. While all methods output plausible right views, the Deep3D image formation model without smoothness constraints does not produce valid disparities.}
  \label{fig:deep3d_compar}
  \vspace{-10pt}
\end{figure}

In Table~\ref{tab:kitti_official}, we see that in addition to having poor scaling properties (in terms of both resolution and the number of disparities it can represent), when trained from scratch with the same network architecture as ours, the Deep3D \cite{xie2016deep3d} image formation model performs poorly.
From Fig.~\ref{fig:deep3d_compar} we can see that Deep3D produces plausible image reconstructions but the output disparities are inferior to ours.
Our loss outperforms both the Deep3D baselines and the addition of the left-right consistency check increases performance in all measures.
In Fig. \ref{fig:lr_consistency_results} we illustrate some zoomed in comparisons, clearly showing that the inclusion of the left-right check improves the visual quality of the results. 
Our results are further improved by first pre-training our model with additional training data from the Cityscapes dataset \cite{Cordts2016Cityscapes} containing $22,973$ training stereo pairs captured in various cities across Germany. 
This dataset brings higher resolution, image quality, and variety compared to KITTI, while having a similar setting. 
We cropped the input images to only keep the top 80\% of the image, removing the very reflective car hoods from the input.
Interestingly, our model trained on Cityscapes alone does not perform very well numerically. 
This is likely due to the difference in camera calibration between the two datasets, but there is a clear advantage to fine-tuning on data that is related to the test set.

\paragraph{Eigen Split}
To be able to compare to existing work, we also use the test split of $697$ images as proposed by \cite{eigen2014depth} which covers a total of $29$ scenes. The remaining $32$ scenes contain $23,488$ images from which we keep $22,600$ for training and the rest for evaluation, similarly to \cite{garg2016unsupervised}.
To generate the ground truth depth images, we reproject the 3D points viewed from the velodyne laser into the left input color camera. 
Aside from only producing depth values for less than $5\%$ of the pixels in the input image, errors are also introduced because of the rotation of the Velodyne, the motion of the vehicle and surrounding objects, and also incorrect depth readings due to occlusion at object boundaries. 
To be fair to all methods, we use the same crop as \cite{eigen2014depth} and evaluate at the input image resolution. 
With the exception of Garg~\ea's~\cite{garg2016unsupervised} results, the results of the baseline methods are recomputed by us given the authors's original predictions to ensure that all the scores are directly comparable. 
This produces slightly different numbers than the previously published ones, \eg in the case of \cite{eigen2014depth}, their predictions were evaluated on much smaller depth images ($1/4$ the original size).
For all baseline methods we use bilinear interpolation to resize the predictions to the correct input image size. 

Table~\ref{tab:kitti_eigen} shows quantitative results with some example outputs shown in Fig. \ref{fig:kitti_eigen}.
We see that our algorithm outperforms all other existing methods, including those that are trained with ground truth depth data. 
We again see that pre-training on the Cityscapes dataset improves the results over using KITTI alone.

\begin{figure*}
  \centering
  \resizebox{0.99\textwidth}{!}{
  \newcommand{\turnheightnew}{0.195\columnwidth}

\centering

\begin{tabular}{@{\hskip 1mm}c@{\hskip 1mm}c@{\hskip 1mm}c@{\hskip 1mm}c@{\hskip 1mm}c@{\hskip 1mm}c@{}}

\Large{Input} & \Large{GT} & \Large{Karsch \ea\cite{karsch2014depth}} & \Large{Liu \ea\cite{liu2014discrete}} & \Large{Laina \ea\cite{laina2016deeper}} & \Large{Ours} \\

\includegraphics[height=\turnheightnew]{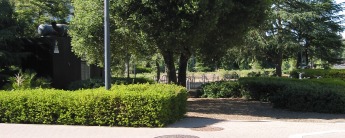} &
\includegraphics[height=\turnheightnew]{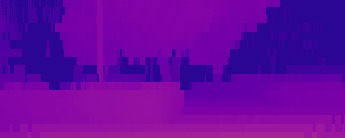} &
\includegraphics[height=\turnheightnew]{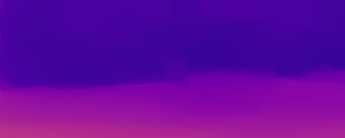} &
\includegraphics[height=\turnheightnew]{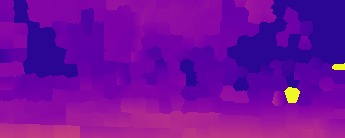} &
\includegraphics[height=\turnheightnew]{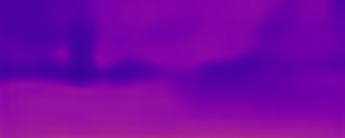} &
\includegraphics[height=\turnheightnew]{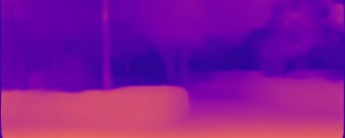} \\

\includegraphics[height=\turnheightnew]{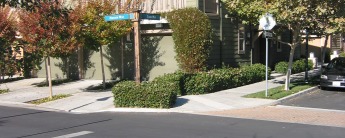} &
\includegraphics[height=\turnheightnew]{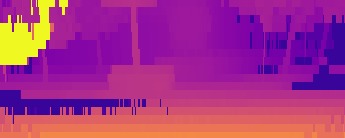} &
\includegraphics[height=\turnheightnew]{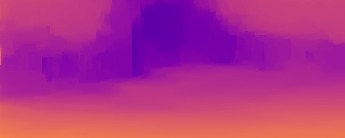} &
\includegraphics[height=\turnheightnew]{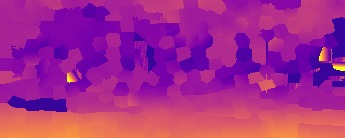} &
\includegraphics[height=\turnheightnew]{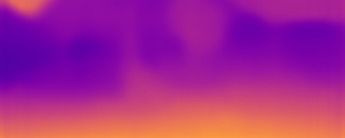} &
\includegraphics[height=\turnheightnew]{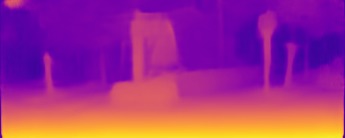} \\

\end{tabular} }
  \vspace{5pt}
  \caption{Our method achieves superior qualitative results on Make3D despite being trained on a different dataset (Cityscapes).}
  \label{fig:make3d_qual}
  \vspace{-10pt}
\end{figure*}

\subsection{Stereo}\label{sec:Stereo}
\begin{figure}[!ht]
  \centering
  \resizebox{\linewidth}{!}{
	\newcommand{\turnheightnew}{0.195\columnwidth}

\centering

\begin{tabular}{@{\hskip 1mm}c@{\hskip 1mm}c@{\hskip 1mm}c@{\hskip 1mm}c@{}}

{\rotatebox{90}{\hspace{5pt}Input left}} &
\includegraphics[height=\turnheightnew]{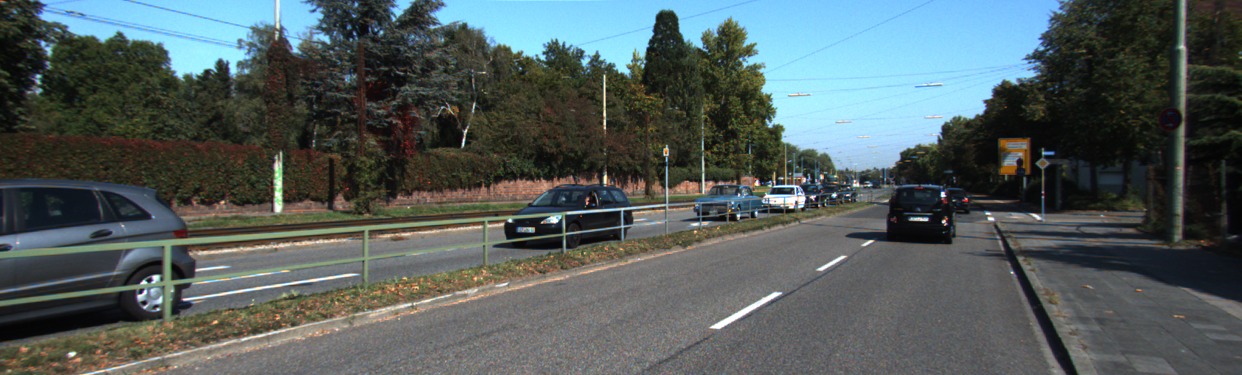} &
\includegraphics[height=\turnheightnew]{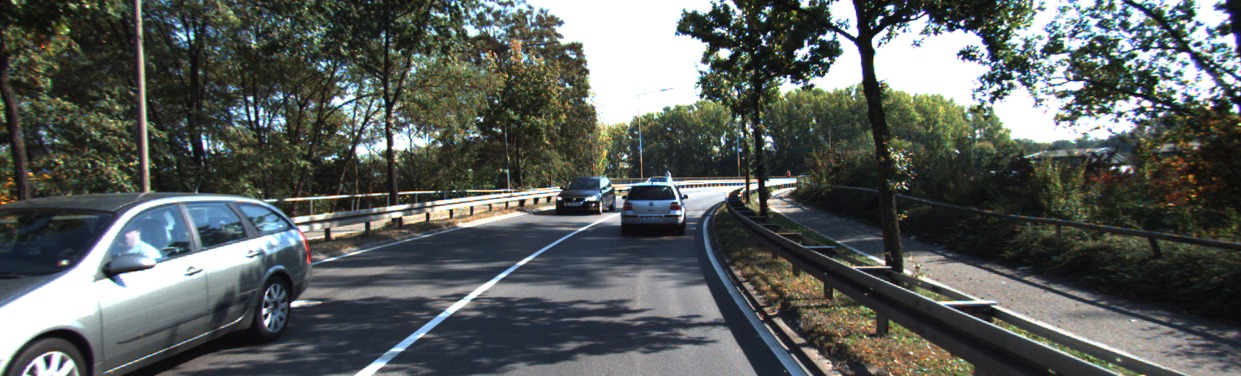} &
\includegraphics[height=\turnheightnew]{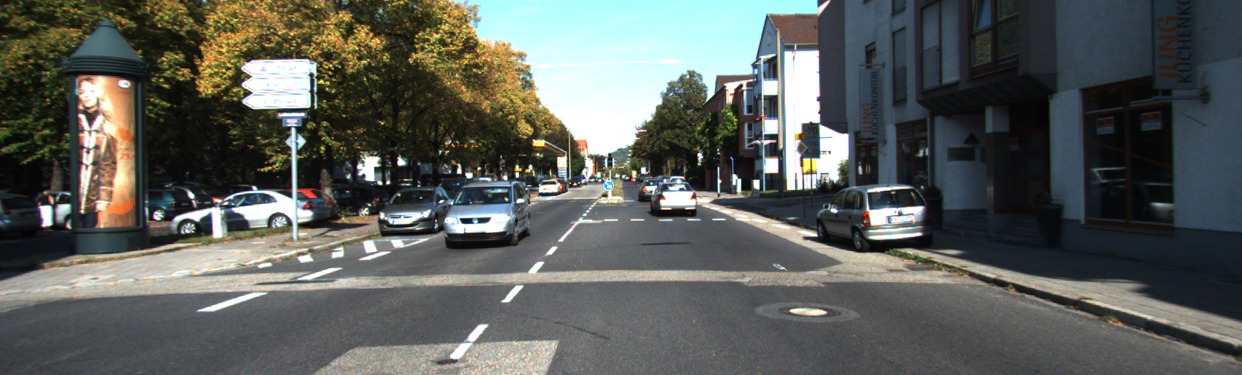} \\

{\rotatebox{90}{\hspace{5pt}Stereo}} &
\includegraphics[height=\turnheightnew]{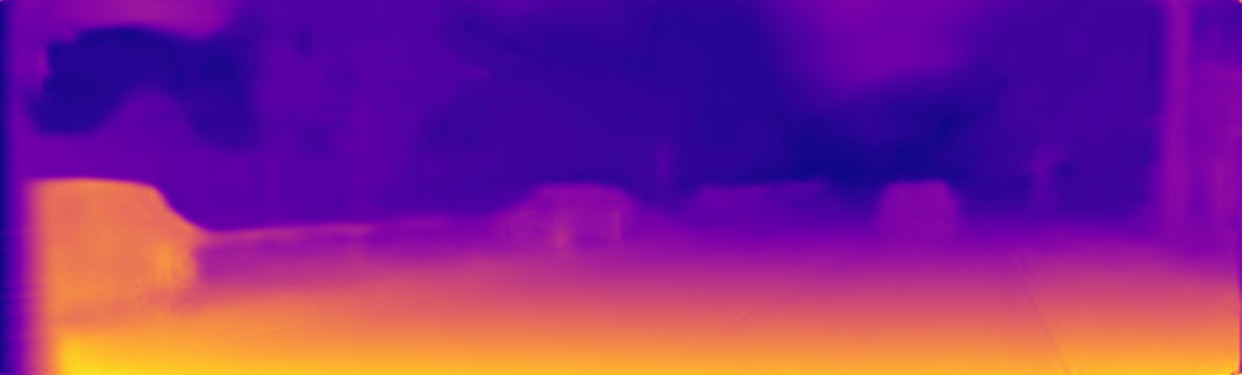} &
\includegraphics[height=\turnheightnew]{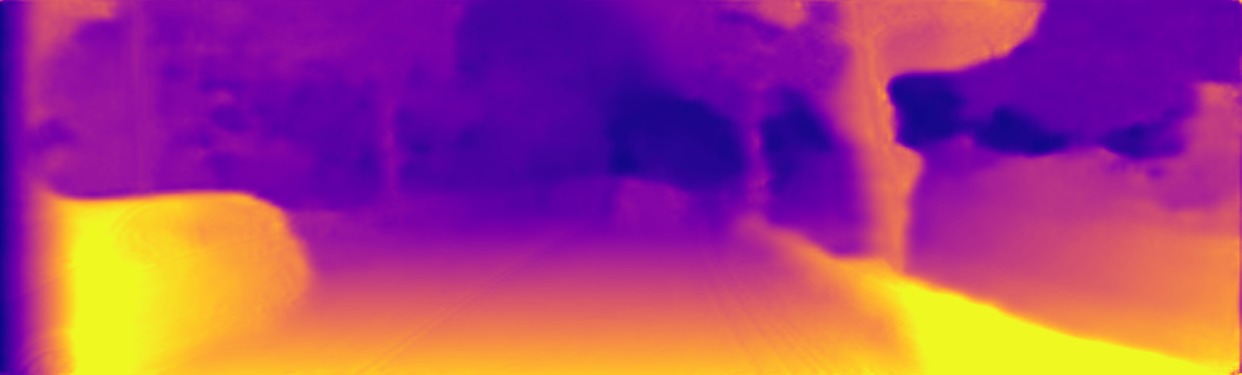} &
\includegraphics[height=\turnheightnew]{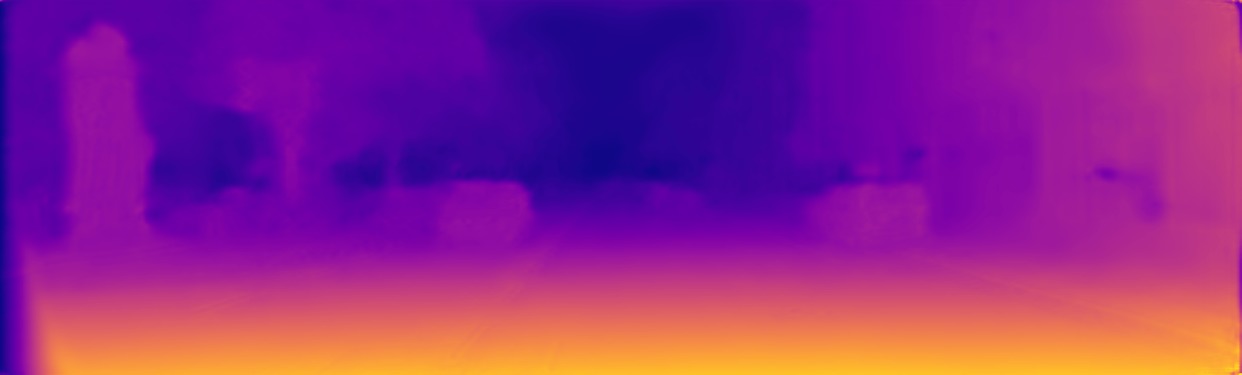} \\

{\rotatebox{90}{\hspace{5pt}Mono}} &
\includegraphics[height=\turnheightnew]{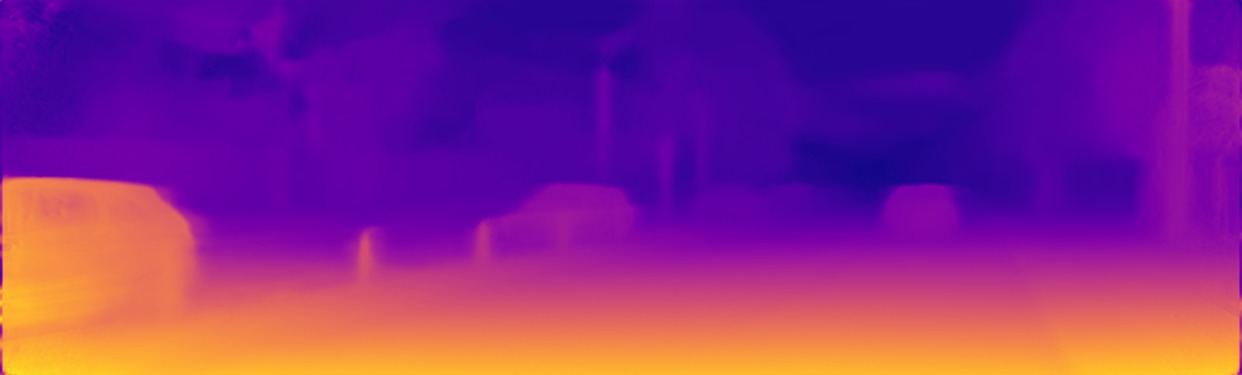} &
\includegraphics[height=\turnheightnew]{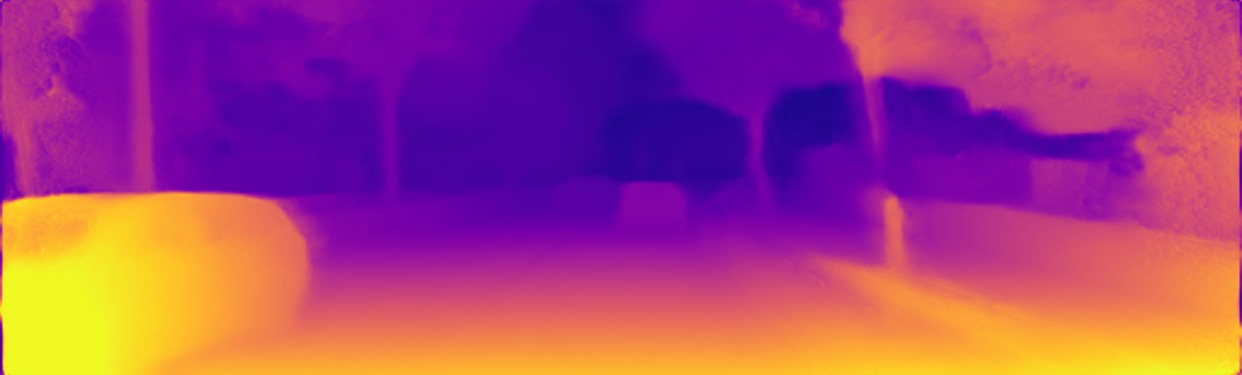} &
\includegraphics[height=\turnheightnew]{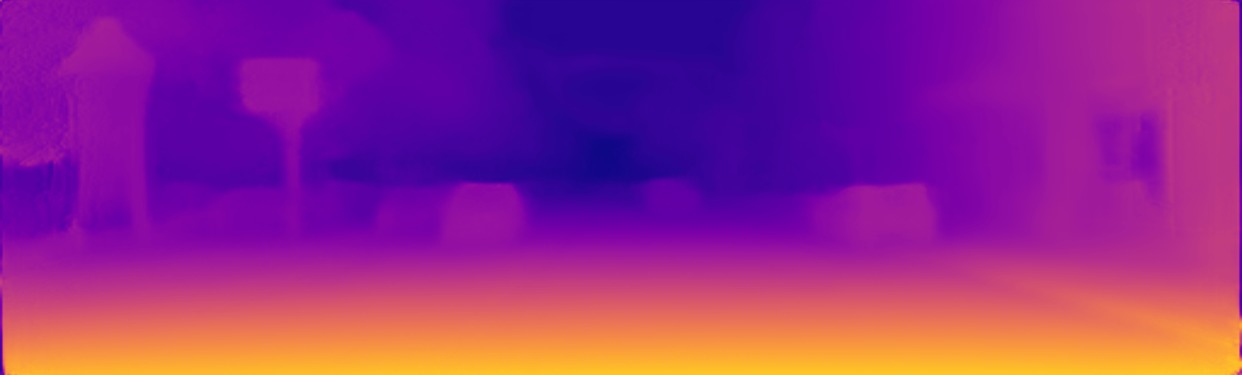} \\

\end{tabular} }
    \vspace{3pt}
  \caption{Our stereo results. While the stereo disparity maps contains more detail, our monocular results are comparable.}
  \label{fig:stereo_results}
  \vspace{-8pt}
\end{figure}

We also implemented a stereo version of our model, see Fig.~\ref{fig:stereo_results}, where the network's input is the concatenation of both left and right views.
Perhaps unsurprisingly, the stereo models outperforms our monocular network on every single metric, especially on the {\it D1-all} disparity measure, as can be seen in Table~\ref{tab:kitti_official}. This model was only trained for 12 epochs as it becomes unstable if trained for longer.

\subsection{Make3D} 
To illustrate that our method can generalize to other datasets, here we compare to several fully supervised methods on the Make3D test set of \cite{saxena2009make3d}. 
Make3D consists of only RGB/Depth pairs and no stereo images, thus our method cannot train on this data.
We use our network trained only on the Cityscapes dataset and despite the dissimilarities in the datasets, both in content and camera parameters, we still achieve reasonable results, even beating \cite{karsch2014depth} on one metric and \cite{liu2014discrete} on three.
Due to the different aspect ratio of the Make3D dataset we evaluate on a central crop of the images.
In Table \ref{tab:make3d_tab}, we compare our output to the similarly cropped results of the other methods.
As in the case of the KITTI dataset, these results would likely be improved with more relevant training data. 
A qualitative comparison to some of the related methods is shown in Fig. \ref{fig:make3d_qual}.
While our numerical results are not as good as the baselines, qualitatively, we compare favorably to the supervised competition.

\begin{table}
  \centering
  \resizebox{0.85\linewidth}{!}{
 \begin{tabular}{|l|c|c|c|c|}
 \hline
 Method & Sq Rel & Abs Rel & RMSE & $\text{log}_{10}$ \\
 \hline
 Train set mean\mbox{*} & 15.517 & 0.893 & 11.542 & 0.223 \\ 
 Karsch \ea\cite{karsch2014depth}\mbox{*} & 4.894 & 0.417 & 8.172 & 0.144 \\ 
 Liu \ea\cite{liu2014discrete}\mbox{*} & 6.625 & 0.462 & 9.972 & 0.161 \\ 
 Laina \ea\cite{laina2016deeper} berHu\mbox{*} & \textbf{1.665} & \textbf{0.198} & \textbf{5.461} & \textbf{0.082} \\ \hline
 \textbf{Ours} with Deep3D \cite{xie2016deep3d} & 17.18 & 1.000 & 19.11 & 2.527 \\
 \textbf{Ours} & 11.990 & 0.535 & 11.513 & 0.156 \\
 \textbf{Ours pp} & 7.112 & 0.443 & 8.860 & 0.142\\ \hline 
\end{tabular}
}
  \vspace{5pt}
  \caption{Results on the Make3D dataset \cite{saxena2009make3d}.
  All methods marked with an \mbox{*} are supervised and use ground truth depth data from the Make3D training set.
  Using the standard C1 metric, errors are only computed where depth is less than 70 meters in a central image crop.}
    \label{tab:make3d_tab}
\end{table}

\subsection{Generalizing to Other Datasets}
\begin{figure}[!hb]
  \centering
  \vspace{2pt}
  \resizebox{0.95\linewidth}{!}{
 \input{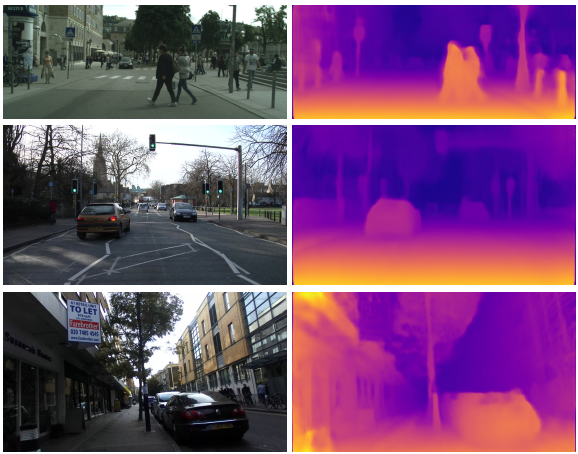} }
\vspace{3pt}
  \caption{Qualitative results on Cityscapes, CamVid, and our own urban dataset captured on foot. For more results please see our video.}
  \label{fig:others}
\end{figure}

Finally, we illustrate some further examples of our model generalizing to other datasets in Figure \ref{fig:others}.
Using the model only trained on Cityscapes \cite{Cordts2016Cityscapes}, we tested on the CamVid driving dataset \cite{brostow2009semantic}. 
In the accompanying video and the supplementary material we can see that despite the differences in location, image characteristics, and camera calibration, our model still produces visually plausible depths. 
We also captured a $60,000$ frame dataset, at 10 frames per second, taken in an urban environment with a wide angle consumer 1080p stereo camera. 
Finetuning the Cityscapes pre-trained model on this dataset produces visually convincing depth images for a test set that was captured with the same camera on a different day, please see the video in the supplementary material for more results.

\subsection{Limitations}
Even though both our left-right consistency check and post-processing improve the quality of the results, there are still some artifacts visible at occlusion boundaries due to the pixels in the occlusion region not being visible in both images. 
Explicitly reasoning about occlusion during training \cite{hoiem2007recovering, humayun_CVPR_2011_occlusions} could improve these issues.
It is worth noting that depending how large the baseline between the camera and the depth sensor, fully supervised approaches also do not always have valid depth for all pixels.

Our method requires rectified and temporally aligned stereo pairs during training, which means that it is currently not possible to use existing single-view datasets for training purposes \eg \cite{Silberman:ECCV12}. 
However, it is possible to fine-tune our model on application specific ground truth depth data.

Finally, our method mainly relies on the image reconstruction term, meaning that specular~\cite{godard2015multi} and transparent surfaces will produce inconsistent depths. 
This could be improved with more sophisticated similarity measures \cite{vzbontar2016stereo}.

%
%
\section{Conclusion}
We have presented an unsupervised deep neural network for single image depth estimation. 
Instead of using aligned ground truth depth data, which is both rare and costly, we exploit the ease with which binocular stereo data can be captured.
Our novel loss function enforces consistency between the predicted depth maps from each camera view during training, improving predictions. 
Our results are superior to fully supervised baselines, which is encouraging for future research that does not require expensive to capture ground truth depth. 
We have also shown that our model can generalize to unseen datasets and still produce visually plausible depth maps.

In future work, we would like to extend our model to videos.
While our current depth estimates are performed independently per frame, adding temporal consistency ~\cite{karsch2014depth} would likely improve results.
It would also be interesting to investigate sparse input as an alternative training signal \cite{zoran2015learning, chen2016single}. 
Finally, while our model estimates per pixel depth, it would be interesting to also predict the full occupancy of the scene \cite{FirmanCVPR2016}.

%
%
\small{
\vspace{8pt}
\noindent\textbf{Acknowledgments} We would like to thank David Eigen, Ravi Garg, Iro Laina and Fayao Liu for providing data and code to recreate the baseline algorithms. We also thank Stephan Garbin for his lua skills and Peter Hedman for his \LaTeX~magic. We are grateful for EPSRC funding for the EngD Centre EP/G037159/1, and for projects EP/K015664/1 and EP/K023578/1.
}

%
%
{\small 
\bibliographystyle{ieee}
\bibliography{mono_depth}
}

\clearpage


\section*{Supplementary material}

# Unsupervised Monocular Depth Estimation with Left-Right Consistency

# Supplementary Material

## 1. Model architecture

| "Encoder" | | | | | | | "Decoder" | | | | | | |
|---|---|---|---|---|---|---|---|---|---|---|---|---|---|
| layer | k | s | chns | in | out | input | layer | k | s | chns | in | out | input |
| conv1 | 7 | 2 | 3/32 | 1 | 2 | left | upconv7 | 3 | 2 | 512/512 | 128 | 64 | conv7b |
| conv1b | 7 | 1 | 32/32 | 2 | 2 | conv1 | iconv7 | 3 | 1 | 1024/512 | 64 | 64 | upconv7+conv6b |
| conv2 | 5 | 2 | 32/64 | 2 | 4 | conv1b | upconv6 | 3 | 2 | 512/512 | 64 | 32 | iconv7 |
| conv2b | 5 | 1 | 64/64 | 4 | 4 | conv2 | iconv6 | 3 | 1 | 1024/512 | 32 | 32 | upconv6+conv5b |
| conv3 | 3 | 2 | 64/128 | 4 | 8 | conv2b | upconv5 | 3 | 2 | 512/256 | 32 | 16 | iconv6 |
| conv3b | 3 | 1 | 128/128 | 8 | 8 | conv3 | iconv5 | 3 | 1 | 512/256 | 16 | 16 | upconv5+conv4b |
| conv4 | 3 | 2 | 128/256 | 8 | 16 | conv3b | upconv4 | 3 | 2 | 256/128 | 16 | 8 | iconv5 |
| conv4b | 3 | 1 | 256/256 | 16 | 16 | conv4 | iconv4 | 3 | 1 | 128/128 | 8 | 8 | upconv4+conv3b |
| conv5 | 3 | 2 | 256/512 | 16 | 32 | conv4b | disp4 | 3 | 1 | 128/2 | 8 | 8 | iconv4 |
| conv5b | 3 | 1 | 512/512 | 32 | 32 | conv5 | upconv3 | 3 | 2 | 128/64 | 8 | 4 | iconv4 |
| conv6 | 3 | 2 | 512/512 | 32 | 64 | conv5b | iconv3 | 3 | 1 | 130/64 | 4 | 4 | upconv3+conv2b+disp4* |
| conv6b | 3 | 1 | 512/512 | 64 | 64 | conv6 | disp3 | 3 | 1 | 64/2 | 4 | 4 | iconv3 |
| conv7 | 3 | 2 | 512/512 | 64 | 128 | conv6b | upconv2 | 3 | 2 | 64/32 | 4 | 2 | iconv3 |
| conv7b | 3 | 1 | 512/512 | 128 | 128 | conv7 | iconv2 | 3 | 1 | 66/32 | 2 | 2 | upconv2+conv1b+disp3* |
| | | | | | | | disp2 | 3 | 1 | 32/2 | 2 | 2 | iconv2 |
| | | | | | | | upconv1 | 3 | 2 | 32/16 | 2 | 1 | iconv2 |
| | | | | | | | iconv1 | 3 | 1 | 18/16 | 1 | 1 | upconv1+disp2* |
| | | | | | | | disp1 | 3 | 1 | 16/2 | 1 | 1 | iconv1 |

Table 1: Our network architecture, where **k** is the kernel size, **s** the stride, **chns** the number of input and output channels for each layer, **in**put and **out**put is the downscaling factor for each layer relative to the input image, and **input** corresponds to the input of each layer where + is a concatenation and ∗ is a 2× upsampling of the layer.

## 2. Post-Processing

The post-processed disparity map corresponds to the per-pixel weighted sum of two components: $d^l$ the disparity of the input image, $d'''^l$ the flipped disparity of the flipped input image.
We define the per-pixel weight map $w^l$ for $d^l$ as

$$w^l(i,j) = \begin{cases} 1 & \text{if } j \leq 0.1 \\ 0.5 & \text{if } j > 0.2 \\ 5*(0.2-i)+0.5 & \text{else,} \end{cases}$$

where $i,j$ are normalized pixel coordinates, and the weight map $w'^l$ for $d'''^l$ is obtained by horizontally flipping $w^l$.
The final disparity is calculated as,
$$d = d^l w^l + d'''^l w'^l.$$

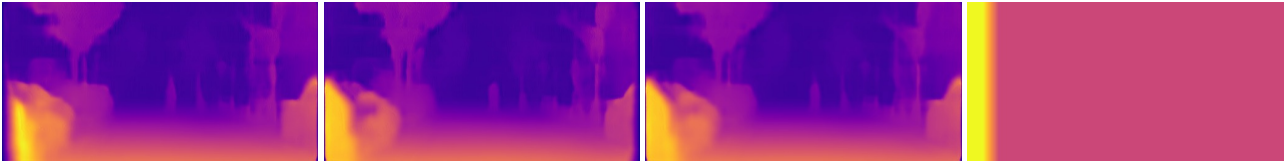

Figure 1: Example of a post-processed disparity map. From left to right: The disparities $d^l$, $d'''^l$, $d$, and the weight map $w^l$.

## 3. Deep3D Smoothness Loss

In the main paper we also compared to our enhanced version of the Deep3D [1] image formation model that includes smoothness constraints on the disparities. Deep3D outputs an intensity image as a weighted sum of offset copies of the input image. The weights $w_i$, seen in Fig. 2a, can be seen as a discrete probability distribution over the disparities for each pixel, as they sum to one. Thus, smoothness constraints cannot be applied directly onto these distributions. However, we see (in Fig. 2c) that if the probability mass is concentrated into one disparity, *i.e.* $\max(w) \approx 1$, then the sum of the cumulative sum of the weights is equal to the position of the maximum. To encourage the network to concentrate probability at single disparities, we added a cost,

$$C_{max} = \frac{1}{N}\|\max(w_i) - 1\|^2, \qquad (1)$$

and its associated weight $\alpha_{max} = 0.02$. Assuming the maximum of each distribution is one, such as in Fig. 2b, meaning the network only picks one disparity per pixel, we can see that the (sum of the) cumulative sum $\mathrm{cs}(w_i)$ of the distribution (Fig. 2c) directly relates to the location of the maximum disparity:

$$d = \operatorname*{argmax}_i(w_i) = n - \sum_i^n \mathrm{cs}(w_i), \qquad (2)$$

where $n$ is the maximum number of disparities.
In the example presented in Fig. 2, we can see that the maximum is located at disparity 3, and that Equation 2 gives us $d = 8 - 6 = 3$. We use this observation to build our smoothness constraint for the Deep3D image formation model.
We can then directly apply smoothness constraints on the gradients of the cumulative sums of the weights at each pixel, so

$$C_{ds} = \frac{1}{N} \sum_{i,j} |\partial_x \mathrm{cs}(w)_{ij}| + |\partial_y \mathrm{cs}(w)_{ij}|, \qquad (3)$$

and its associated weight $\alpha_{ds} = 0.1$.

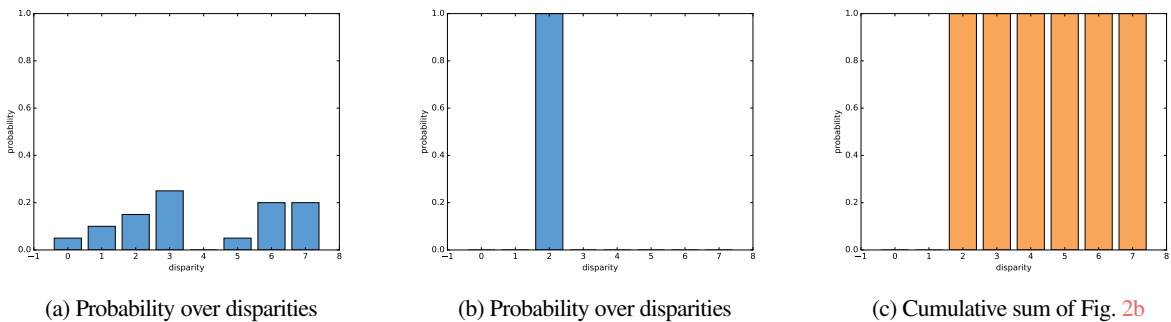

(a) Probability over disparities  (b) Probability over disparities  (c) Cumulative sum of Fig. 2b

Figure 2: Deep3Ds per-pixel disparity probabilities.

## 4. More KITTI Qualitative Results

In Fig. 3 we show some additional qualitative comparisons for the KITTI dataset using the Eigen split.

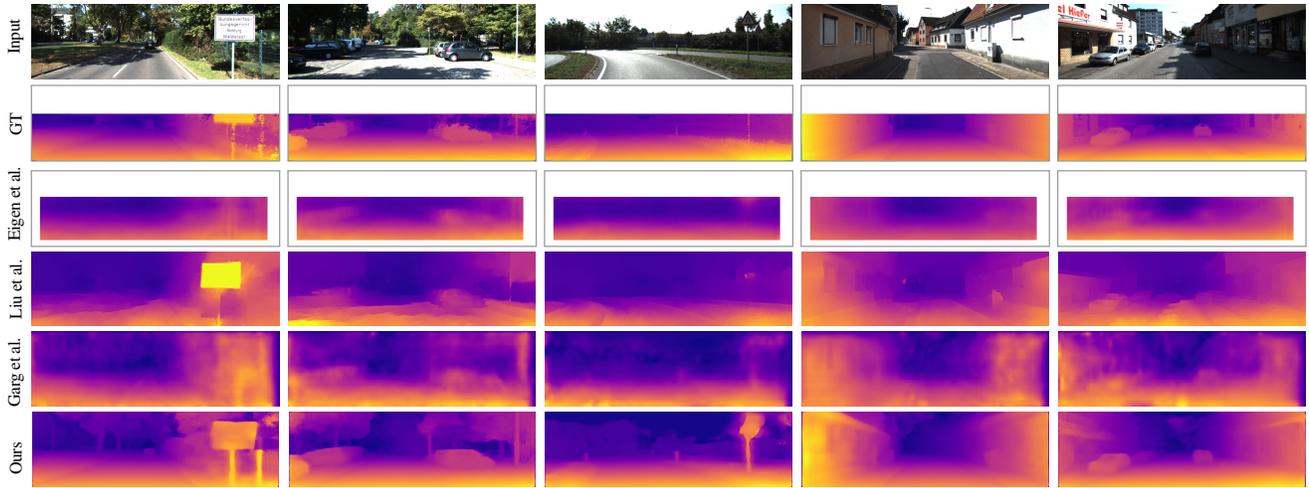

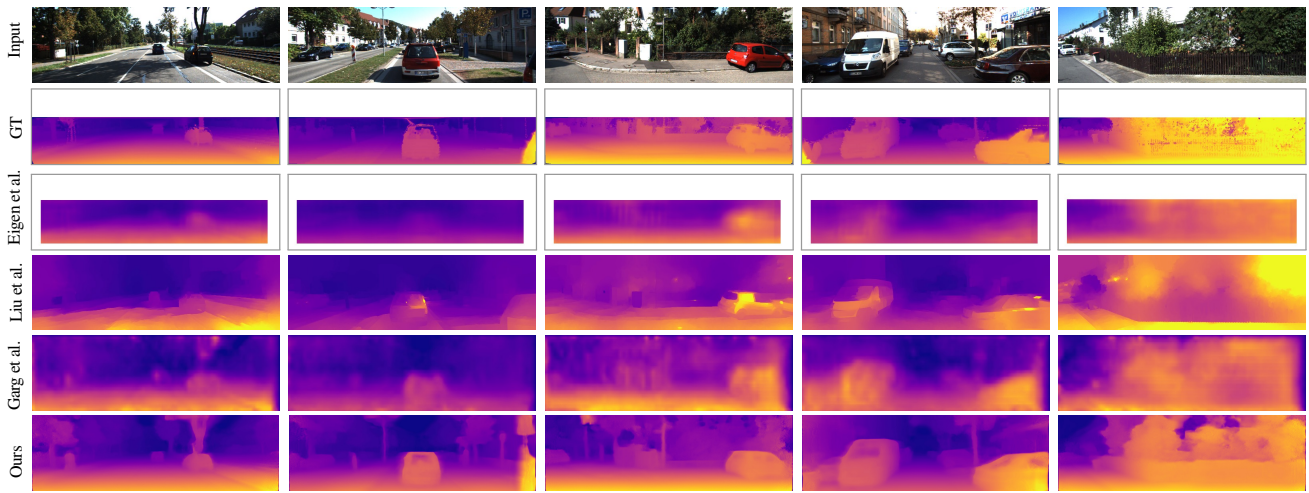

Figure 3: Additional qualitative results on the KITTI Eigen Split. As ground truth velodyne depth is very sparse, we interpolate it for visualization purposes.

## 5. Disparity Error Maps

As we train our model with color augmentations, we can apply the same principle at test time and analyze the distribution of the results. We applied 50 random augmentations to each test images and show the standard deviation of the disparities per pixel (see Figure 4). We can see that the network gets confused with close-by objects, texture-less regions, and occlusion boundaries. Interestingly, one test image was captured in a tunnel, resulting in a very dark image. Our network clearly shows high uncertainty for this sample as it is very different from the rest of the training set.

## References

[1] J. Xie, R. Girshick, and A. Farhadi. Deep3d: Fully automatic 2d-to-3d video conversion with deep convolutional neural networks. In *ECCV*, 2016. 2

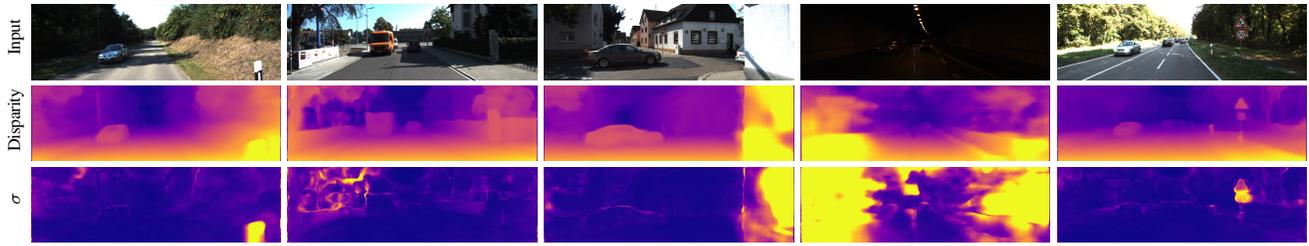

Figure 4: Uncertainty of our model on the KITTI dataset. From top to bottom: input image, predicted disparity, and standard deviation of multiple different augmentations. We can see that there is uncertainty in low texture regions and at occlusion boundaries.



\end{document}